\documentclass[runningheads]{llncs}
\pdfoutput=1
 
\usepackage{eccv}

\usepackage{eccvabbrv}

\usepackage{graphicx}
\usepackage{booktabs}
\newcommand{\cmark}{\ding{51}}
\newcommand{\xmark}{\ding{55}}
\usepackage{pifont}

\usepackage[accsupp]{axessibility}  

\usepackage{colortbl}
\usepackage[table]{xcolor}
\usepackage{makecell}
\usepackage{subcaption}
\usepackage{booktabs}
\usepackage[accsupp]{axessibility}
\usepackage{xcolor}
\usepackage{listings}
\usepackage{float}
\usepackage{amsmath}
\usepackage{enumitem}
\usepackage{array}

\definecolor{promptbg}{HTML}{F4F4F4}
\definecolor{promptframe}{HTML}{BBBBBB}

\lstdefinestyle{PS}{
  backgroundcolor   = \color{promptbg},
  basicstyle        = \ttfamily\fontsize{5.1}{7.0}\selectfont,
  breaklines        = true,
  breakatwhitespace = false,
  columns           = fullflexible,
  keepspaces        = true,
  frame             = single,
  framesep          = 3pt,
  rulecolor         = \color{promptframe},
  aboveskip         = 0pt,
  belowskip         = 0pt,
  xleftmargin       = 0pt,
  xrightmargin      = 0pt,
  lineskip          = 0pt,
}

\usepackage{graphicx}
\usepackage{booktabs}
\usepackage[accsupp]{axessibility}
\usepackage{tcolorbox}
\usepackage{hyperref}
\usepackage{orcidlink}
\usepackage{multirow}

\usepackage{hyperref}

\usepackage{orcidlink}

\begin{document}

\title{\textbf{\textsc{SpatiO}}: Adaptive Test-Time Orchestration of Vision-Language Agents for Spatial Reasoning} 

\titlerunning{\textbf{\textsc{SpatiO}}}

\author{Chan Yeong Hwang\inst{1}, 
Miso Choi\inst{1}, Sunghyun On\inst{3}, \\ Jinkyu Kim\inst{1,2}$^\dag$, Jungbeom Lee\inst{1}$^\dag$ }

\authorrunning{C. Y. Hwang et al.}

\institute{
Korea University \quad
Kakao Mobility Corp. \quad
Handong Global University\\
\email{\{flaxinger, miso8070, jinkyukim, jbeomlee\}@korea.ac.kr}\\
\email{22100457@handong.ac.kr}\\[0.5em]
$^\dag$Co-corresponding authors.
}

\maketitle

\vspace{-1mm}
\begin{abstract}
Understanding visual scenes requires not only recognizing objects but also reasoning about their spatial relationships. 
Unlike general vision-language tasks, spatial reasoning requires integrating multiple inductive biases, such as 2D appearance cues, depth signals, and geometric constraints, whose reliability varies across contexts. 
This suggests that effective spatial reasoning requires \emph{spatial adaptability}: the ability to flexibly coordinate different reasoning strategies depending on the input.
However, most existing approaches rely on a single reasoning pipeline that implicitly learns a fixed spatial prior, limiting their ability to adapt under distribution changes. 
Multi-agent systems offer a promising alternative by aggregating diverse reasoning trajectories, but prior attempts in spatial reasoning primarily employ homogeneous agents, restricting the diversity of inductive biases they can leverage.
In this work, we introduce \textbf{\textsc{SpatiO}}, a heterogeneous multi-agent framework for spatial reasoning that coordinates multiple vision-language specialists with complementary inductive biases. 
To enable effective collaboration, we propose \textbf{Test-Time Orchestration (TTO)}, an calibration mechanism that dynamically evaluates and reweights agents based on their observed reliability during inference, without modifying model parameters.
Extensive experiments on diverse spatial reasoning benchmarks, including 3DSRBench, STVQA-7k, CV-Bench, and Omni3D-Bench, demonstrate that \textsc{SpatiO} consistently improves spatial reasoning performance over both closed-source and open-source baselines. The project page is available at \url{https://cy-h1329.github.io/spatio/}.
\end{abstract}
\section{Introduction}
\label{sec:intro}

\begin{figure}[!t]
  \centering
  \includegraphics[width=\linewidth]{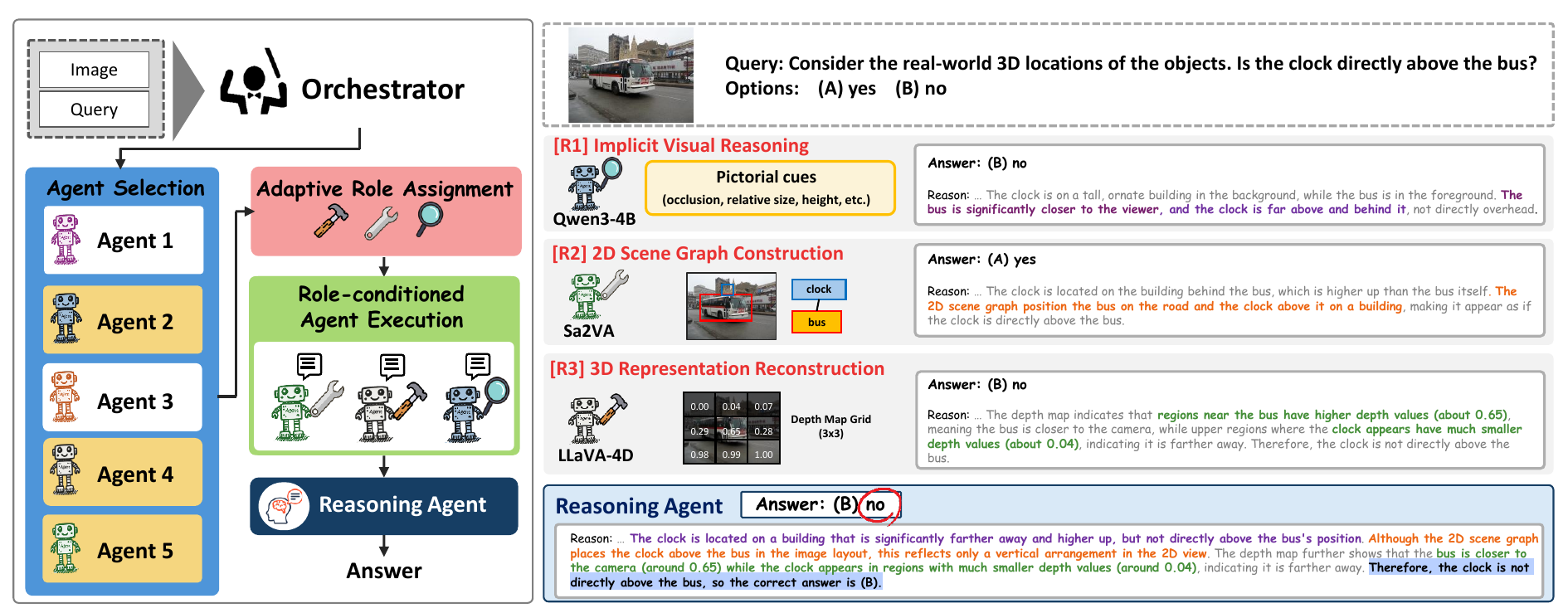}
  \caption{\textbf{Overall framework illustration of \textsc{SpatiO}.} Multiple heterogeneous vision-language agents operate as specialists under different roles, and their outputs are aggregated to produce the final spatial reasoning result.}
  \label{fig:teaser}
\end{figure}

\vspace{-0.5mm}

Understanding visual scenes requires more than recognizing objects and attributes; it demands reasoning about how entities are arranged in space, how they relate to one another, and how these relationships change across viewpoints~\cite{ma20253dsrbench,tong2024cambrian,stogiannidismind,xue2025reasoning}.
This capability, commonly referred to as \textbf{spatial reasoning}, is fundamental for visual intelligence.
It is becoming increasingly essential as artificial systems aim to interpret, navigate, and interact with real-world environments, ranging from embodied agents, robotics, to augmented and virtual reality, 3D scene understanding, and physical world modeling.
Unlike general visual tasks such as image captioning or visual question answering, spatial reasoning requires integrating heterogeneous sources of information, such as 2D appearance cues, depth signals, spatial relations, and implicit geometric priors \cite{zhang2025open3d}.
Crucially, the reliability of these cues is highly context-dependent.
In canonical views, appearance cues are often sufficient, whereas geometric constraints become critical under occlusion or clutter.
As a reTsult, effective performance of spatial reasoning hinges on \textbf{spatial adaptability}, the ability to flexibly coordinate different strategies in response to the specific conditions of a given task.

The predominant approach for spatial reasoning has been large-scale supervision: training on billions of synthetic spatial VQA examples \cite{chen2024spatialvlm}, or incorporating structured representations of location and orientation derived from 3D data~\cite{ma2025spatialllm}.
These models achieve competitive performance by internalizing powerful spatial priors tailored to their training distributions, effectively functioning as specialists for particular categories of spatial queries.
Subsequent approaches~\cite{batra2025spatialthinker,ma2025spatialreasoner}
further enhance reasoning by introducing explicit structural priors and reinforcement learning, improving data efficiency and inference quality within a single reasoning pipeline. 
Despite these advances, existing approaches share a common limitation: they realize spatial reasoning through a single-model reasoning pipeline that implicitly internalizes a fixed spatial prior tied to its training distribution. In practice, this induces strong but narrow inductive biases: the model behaves as a specialist for the configurations and query types it has frequently seen, while generalizing poorly to out-of-distribution layouts or viewpoints~\cite{ma2025spatialllm,ma20253dsrbench}. Since spatial
reasoning is inherently multifaceted, spanning depth estimation,
orientation inference, and multi-object relational reasoning, any single model tuned to a fixed spatial prior inevitably overfits certain facets of this space while neglecting others. As also observed in recent spatial benchmarks, different models excel on different spatial categories, with decorrelated failure modes across depth, orientation, and relational tasks~\cite{ma20253dsrbench,openai_gpt52_2025}. Even methods that introduce intermediate reasoning steps remain confined to a fixed architectural perspective, limiting their ability to integrate
complementary spatial cues from structurally diverse models.

However, spatial reasoning fundamentally involves multiple inductive biases, whose reliability is inherently context-dependent rather than fixed.
This raises a fundamental question:

\begin{tcolorbox}[
  colback=gray!8,
  colframe=gray!40,
  boxrule=0.5pt,
  arc=2pt,
  left=6pt,
  right=6pt,
  top=6pt,
  bottom=6pt
]  
How can a system dynamically determine which reasoning strategies to trust for a given input and modulate their contributions accordingly at test time?
\end{tcolorbox}

\noindent With this perspective, we reframe spatial reasoning as a coordination problem, where multiple complementary reasoning strategies are adaptively combined at test time rather than optimized within a single static model.

Prior work~\cite{wang2024mixture,yang2025multi} has shown that heterogeneous agents with decorrelated failure modes improve robustness under distribution shift, suggesting that separating inductive biases across specialists offers a principled path toward adaptability. However, existing attempt~\cite{marsili2025visual} to apply this heterogeneous-agent paradigm to spatial reasoning remain limited: replicating a homogeneous backbone or restricting agents to peripheral roles fails to address conditional reliability across reasoning strategies. As a result, the system neither maintains role-dependent estimates of when a given strategy is trustworthy, nor down-weights specialists when their outputs are noisy. Because multi-step pipelines amplify upstream errors, effective orchestration must dynamically reweight contributions based on context-specific reliability, motivating a framework for reliability-aware collaboration among complementary VLM specialists.

We introduce \textbf{\textsc{SpatiO}}, a heterogeneous multi-agent framework for spatial reasoning through adaptive test-time coordination. \textsc{SpatiO} assembles a diverse pool of VLMs with distinct architectures, training objectives, and geometric inductive biases, each independently solving the spatial query under a designated reasoning role. We propose a novel \textbf{Test-Time Orchestration (TTO)} mechanism that dynamically reweights agents based on observed reliability, leveraging per-agent confidence scores to continuously update the orchestration strategy, enabling parameter-free adaptation without modifying model weights. By adapting collaboration rather than parameters, \textsc{SpatiO} avoids catastrophic forgetting, incurs minimal overhead, and remains compatible with both open-source and black-box models. Our primary contributions are as follows:

\vspace{-0.9mm}
\begin{enumerate}
    \item We propose \textbf{\textsc{SpatiO}}, a fully heterogeneous, role-based multi-agent framework for spatial reasoning that coordinates complementary pretrained VLMs without any parameter updates.
    \item We formulate \textbf{Test-Time Orchestration (TTO)}, a parameter-free mechanism that dynamically adjusts agent selection and aggregation weights via Bayesian trust modeling and dual EMA filtering.
    \item We empirically show that \textsc{SpatiO} outperforms monolithic VLMs across challenging spatial benchmarks including 3DSRBench, STVQA, CV-Bench and Omni3D-Bench.
\end{enumerate}

\section{Related Work}
\label{sec:related}

\subsection{Spatial Intelligence in Vision-Language Models}
Research on enhancing spatial reasoning in VLMs has progressed from implicit feature learning to explicit geometric modeling. Initial studies primarily utilized large-scale supervision. For example, SpatialVLM~\cite{chen2024spatialvlm} was trained on billions of synthetic spatial visual question answering (VQA) pairs to facilitate quantitative distance estimation, while SpatialLLM~\cite{ma2025spatialllm} transformed 3D spatial data into structured textual representations. Although these data-intensive methods demonstrate strong in-distribution performance, models such as SpatialLadder~\cite{li2025spatialladder} still exhibit noticeable degradation on spatial benchmarks whose task categories and data distributions differ from their synthetic training datasets. This limitation arises from the reliance on distribution-specific spatial heuristics rather than the development of compositional reasoning capabilities.

Later research introduced explicit structural priors to address these limitations. SpatialRGPT~\cite{cheng2024spatialrgpt} integrated 3D region proposals and scene-graph priors directly into VLM attention mechanisms. SpatialThinker~\cite{batra2025spatialthinker} combined scene-graph grounding with online reinforcement learning guided by a multi-objective reward, achieving state-of-the-art results with only 7K training samples. 
Spatial\-Reasoner-R1~\cite{ma2025spatialreasoner} utilized fine-grained direct preference optimization (fDPO) and long chain-of-thought supervision to systematically parse complex spatial configurations. 
Orthogonal to this, RegionVLM~\cite{lee2024toward} adds explicit, user-indicated regional grounding to VLP models without architectural changes, a further inductive bias for pools such as ours; however, all of these approaches realize spatial reasoning through a single, fixed inference pipeline, limiting adaptation to out-of-distribution configurations or novel viewpoints.

\subsection{Multi-Agent Systems for Spatial Reasoning}
The ensembling of multiple large language models (LLMs) has demonstrated significant effectiveness in language reasoning tasks. MoA~\cite{wang2024mixture} and MoSA~\cite{yang2025multi} show that synthesizing diverse reasoning trajectories from heterogeneous models consistently outperforms individual model performance. Notably, \textit{Debate or Vote}~\cite{DebateOrVote} establishes that when agents share the same underlying model, inter-agent debate does not improve expected correctness; any observed gains are attributable to majority voting over identically-biased outputs. This finding underscores that the benefits of multi-agent systems (MAS) are fundamentally dependent on genuine architectural heterogeneity.

In the visual domain, VADAR~\cite{marsili2025visual} outperforms program-synthesis baselines~\cite{ViperGPT,gupta2023visual} but depends on a single GPT-4o backbone with rule-based aggregation. This approach exemplifies the homogeneous failure mode described in \textit{Debate or Vote}, resulting in underperformance compared to a standard single-pass GPT-4o baseline. \textsc{SpatiO} addresses these limitations by pooling vision-language models (VLMs) with decorrelated error distributions and replacing rule-based aggregation with trust-weighted, geometry-aware re-generation.

More recently, GCA~\cite{chen2025geometrically} assigns dual roles of semantic planner and task executor within a single Qwen3-VL-Thinking backbone, thereby improving spatial inference robustness within a unified system. However, this fixed single-agent architecture is unable to leverage the complementary inductive biases that arise from the coordination of multiple heterogeneous specialists.

\vspace{-2mm}
\subsection{Reliability Estimation and Test-Time Adaptation}
\vspace{-2mm}
A primary challenge in multi-agent spatial reasoning is estimating agent reliability under dynamic conditions. MATE~\cite{algazinov2025mate} addresses this via a Beta–Bernoulli trust model updated online, but is designed for abstract cooperative agents without structured geometric evidence or role-dependent behaviors in VLMs.

At a finer granularity, \cite{choi2026truth} shows that context-truthful attention heads, inherited within model lineages, can be selectively amplified to improve contextual grounding, complementary to \textsc{SpatiO}'s agent-level trust estimation.

Test-Time Adaptation (TTA)~\cite{sun2020test,liang2025comprehensive} adapts pretrained models to distribution shifts via parameter-level updates (e.g., entropy minimization), but is restricted to single-model backbones with mutable weights, limiting applicability when coordinating black-box VLMs.

\textsc{SpatiO} integrates both paradigms into a spatially grounded formulation: rather than adapting parameters, test-time adaptation operates at the orchestration level by maintaining a Bayesian trust estimator over agent–role–category triplets, smoothed via dual EMA and used to dynamically reweight heterogeneous specialists. This yields a reliability-aware coordination mechanism tailored to multimodal spatial reasoning, rather than generic agent interaction or single-model adaptation.

\section{Observation}
\vspace{-1mm}
\subsection{Decorrelated Error Patterns across Spatial Specialists}
\label{sec:decorrelated_error}
\vspace{-1mm}
To better understand the limitations of existing spatial reasoning systems, we analyze five representative vision-language models spanning diverse training paradigms, including large-scale synthetic spatial VQA supervision, structured preference alignment, and reinforcement learning with verifiable rewards. The models further differ in their representational assumptions, ranging from dense 2D grounding mechanisms to explicit 3D region-aware attention and scene-graph–based reasoning.
We evaluate these models on three spatial reasoning benchmarks including 3DSRBench~\cite{ma20253dsrbench}, STVQA-7k~\cite{batra2025spatialthinker}, and CV-Bench~\cite{tong2024cambrian}, and reorganize task categories into five general groups: \textsc{Spatial Relation, Counting, Size, Distance \& Depth}, and \textsc{Orientation}. 

\begin{figure}[H]
  \centering
  \includegraphics[width=\linewidth]{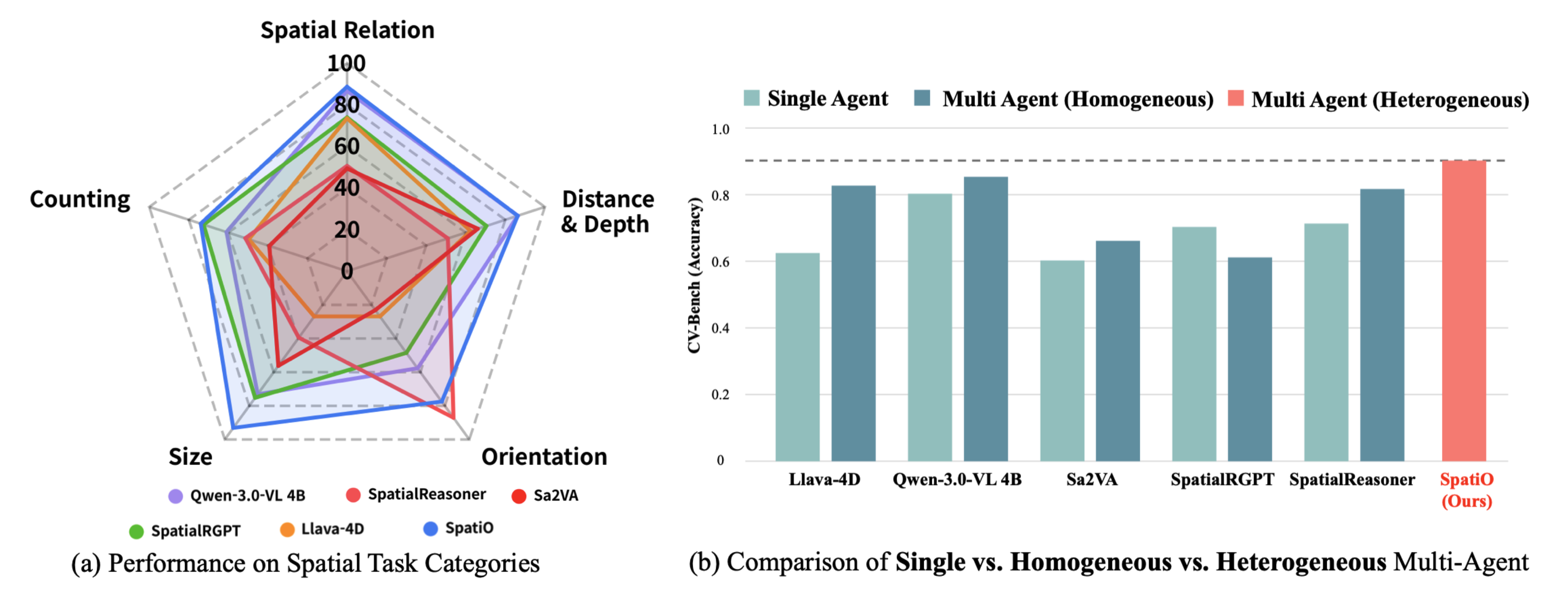}
  \caption{(a) Per-spatial task accuracy of five spatial reasoning specialists (single-agent) and \textsc{Spatio} (Ours) on the reorganized task groups. Each agent occupies a distinct region of the performance space, motivating heterogeneous orchestration over single-backbone selection.} 
  \label{fig:per_task_radar}
\end{figure}

Fig.~\ref{fig:per_task_radar}.(a) presents a radar plot summarizing per-group performance across models.
A consistent pattern emerges: no single model uniformly excels across all spatial groups. Instead, each model exhibits category-specific strengths accompanied by systematic weaknesses. For instance, SpatialReasoner~\cite{ma2025spatialreasoner}, which incorporates explicit multi-stage 3D structural representations, performs strongly on orientation-sensitive queries but shows degradation on distance and depth estimation tasks. In contrast, SpatialRGPT~\cite{cheng2024spatialrgpt}, which leverages 3D scene graphs for data generation and integrates depth-aware features,  demonstrates stronger performance on distance and depth–related queries while underperforming on other relational categories.

We interpret this behavior as evidence that training under fixed distributions encourages models to internalize powerful yet selective spatial priors. Consequently, each model functions as a specialist, resulting in decorrelated error patterns across the model pool.

\vspace{-1mm}

\subsection{Homogeneous vs. Heterogeneous Multi-Agent Systems}
Building on the specialist behaviors identified in Section~\ref{sec:decorrelated_error}, a logical progression is to deploy multiple agents and aggregate their outputs, given that no single model is universally reliable.
Prior work such as VADAR~\cite{marsili2025visual} explores homogeneous multi-agent systems for spatial reasoning, deploying multiple instances of a single backbone. However, as shown by~\cite{DebateOrVote}, such ensembles yield only marginal gains: agents sharing the same architecture exhibit identical inductive biases and failure modes, causing aggregation to amplify shared errors rather than correct them.
To verify this (Fig.~\ref{fig:per_task_radar}b), we deploy three instances of the same backbone with distinct role-based prompts. While aggregation improves over the single-agent baseline, confirming that prompt-level diversity introduces variation in reasoning trajectories, the gains remain modest due to shared architecture and training-induced biases.
This motivates \textbf{architectural heterogeneity}. Combining distinct spatial specialists (e.g., Qwen3-VL-4B and SpatialReasoner) yields substantially larger gains across benchmarks, as each model contributes complementary geometric perspectives inaccessible within any single backbone.

\section{Method}
\subsection{Overall Framework}
\label{sec:spatial_mas}

\begin{figure*}[t]
  \centering
  \includegraphics[width=\textwidth]{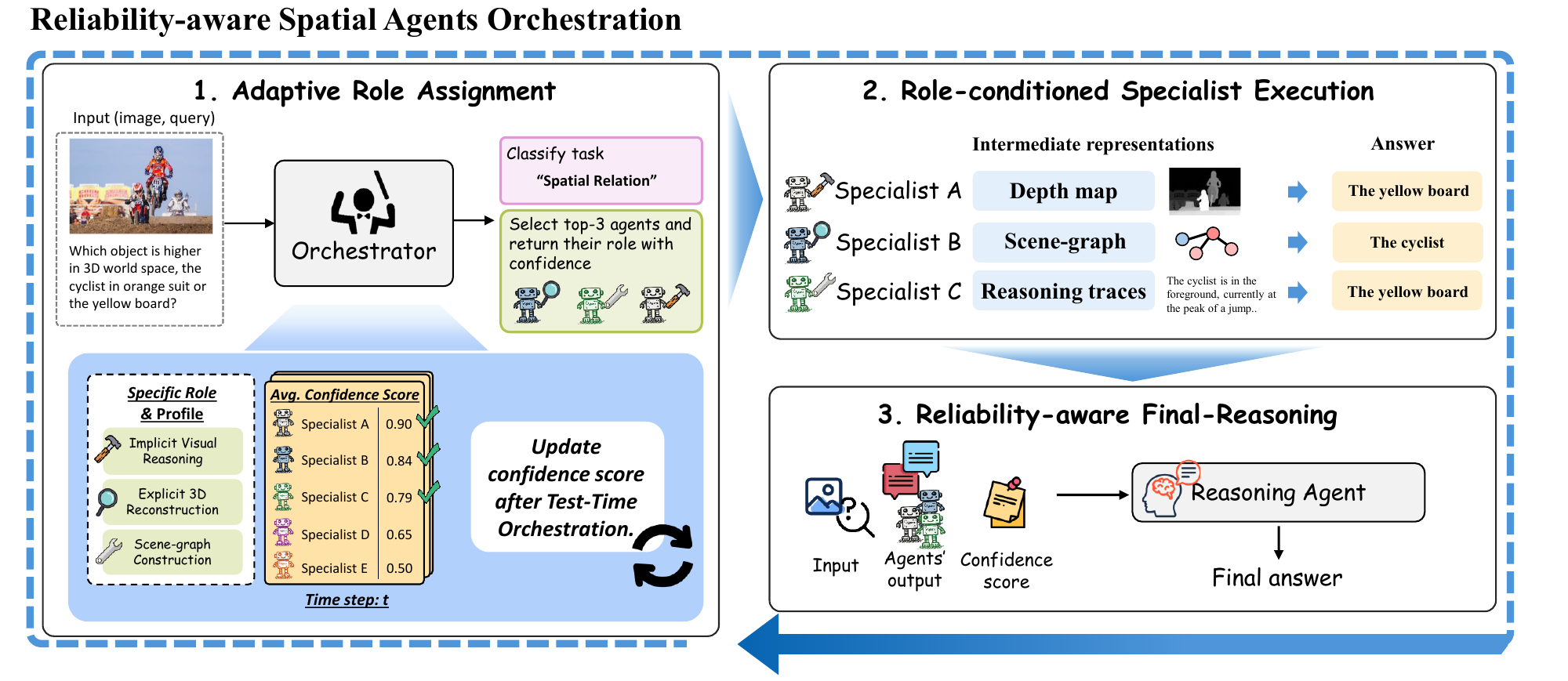}
  \caption{Overview of the \textsc{SpatiO} framework. \textit{(1)} The head agent classifies the query and selects the Top-3 agents with role assignments and trust weights. \textit{(2)} Each agent independently reasons under its designated role, optionally invoking open-source tools; outputs are stored in shared memory $\mathcal{M}^{(t)}$. \textit{(3)} The reasoning agent performs conditional evidence integration and emits $\hat{y}^{(t)}$; trust scores are updated via Bayesian estimation and dual EMA.\\}
  \label{fig:architecture}
\end{figure*}

As established in Section~\ref{sec:intro}, no single vision-language model (VLM) demonstrates uniform superiority across spatial task categories, and their error distributions are largely uncorrelated (Fig.~\ref{fig:per_task_radar}). \textsc{SpatiO} leverages this complementarity by treating pretrained VLMs as fixed specialist members within a structured team, replacing parameter-level adaptation with orchestration-level optimization. Fig.~\ref{fig:architecture} illustrates the overall pipeline, which comprises three sequential components:
\textit{Query Routing},
\textit{Role-Based Analysis},
and \textit{Weighted Synthesis} (Section~\ref{sec:synthesis}). 
Each component is described in detail in the following subsections.

\subsection{Preliminaries}

\noindent{\textbf{Agent Pool.   }}
\label{sec:agent_pool}
Our candidate pool $\mathcal{A}$ consists of five open-source vision-language models (VLMs) selected to maximize architectural complementarity.
All models can be deployed directly at test time without the need for additional fine-tuning.

\noindent{\textbf{Role Definitions.    }} Each agent in $\mathcal{A}$ can be assigned one of three canonical roles $k \in \mathcal{K}$ via structured system prompts specifying the reasoning strategy, tools, and output format, \textit{prompt-mediated role injection}, enabling any model to assume any role without parameter modification.

\begin{enumerate}
  \item \textbf{Implicit Visual Reasoning}: Performs end-to-end spatial visual question answering (VQA) using the raw 2D image, relying solely on the agent's inherent visual grounding capabilities without external tools.

  \item \textbf{Explicit 3D Reconstruction}: Utilizes DepthPro~\cite{bochkovskii2024depth} for metric monocular depth estimation and SAM2~\cite{ravi2024sam} for instance segmentation, integrating these outputs into a 3D point-cloud representation before generating an answer.

  \item \textbf{Scene-Graph Construction}: Constructs a relational scene graph using DINO-v2~\cite{oquab2023dinov2}, encoding object identities and spatial predicates such as \emph{above}, \emph{left-of}, and \emph{closer-than}. The agent then grounds its answer on this symbolic structure~\cite{cheng2024spatialrgpt,ConceptGraphs}.
\end{enumerate}


\subsection{Spatial Multi-agentic Orchestration}

\begin{table}[h]
\centering
\footnotesize
\caption*{$i$: agent, $k$: role, $c$: category, $t$: time step.}
\setlength{\tabcolsep}{8pt}
\renewcommand{\arraystretch}{1.8}
\resizebox{\textwidth}{!}{%
\begin{tabular}{cp{12.5cm}}
\toprule
\textbf{Symbol} & \textbf{Meaning} \\
\midrule
$s_{i,k,c}^{(t)}$ & Confidence score of agent $i$ \emph{under role $k$} for category $c$ \\
$\bar{s}_{i,c}^{(t)}$ & Role-averaged confidence of agent $i$ for category $c$; used for Top-3 \emph{agent} selection \\
$w_{i,k,c}^{(t)}$ & Normalized reliability \emph{weight} of agent $i$ in role $k$ for category $c$; used in final reasoning \\
\bottomrule
\end{tabular}%
}
\end{table}

\vspace{1mm}
\noindent{\textbf{Adaptive Role Assignment. }}\label{sec:routing}The head agent classifies the incoming pair $(x^{(t)}, I^{(t)})$, consisting of a textual query $x^{(t)}$ and an observed image $I^{(t)}$, into one of $|\mathcal{C}|$ spatial task categories from a unified taxonomy derived from Section \ref{sec:decorrelated_error}, where $\mathcal{C} = \{\textit{Spatial Relation, Counting, Size, Distance \& Depth, Orientation}\}$ with $|\mathcal{C}| = 5$.
$t$ denotes the orchestration time step (Section~\ref{sec:tto}). For each agent $i$, role $k \in \mathcal{K}$, and category $c$, we maintain a confidence score $s_{i,k,c}^{(t)}$, initialized to $0.5$ for unseen categories, and select the top-3 agents via an agent-level aggregated score:
\begin{equation}
    \label{eq:role_avg}
    \bar{s}_{i,c}^{(t)} = \frac{1}{|\mathcal{K}|} \sum_{k \in \mathcal{K}} s_{i,k,c}^{(t)},
\end{equation}

\begin{equation}
    \mathcal{S}_c^{(t)} = \operatorname{Topk}_{i}\big(\bar{s}_{i,c}^{(t)},\, k=3\big), \quad \bar{s}_{i,c}^{(t)} \in [0,1],
\end{equation}
where the selection is performed over all agents $i \in \mathcal{A}$ for the current query category $c$.
For each predefined role $k \in \mathcal{K}$, we assign the agent within $\mathcal{S}_c^{(t)}$ that achieves the highest confidence score $s_{i,k,c}^{(t)}$ (ties are resolved using $\bar{s}_{i,c}^{(t)}$).
We denote by $k_i$ the role assigned to agent $i$ under this selection rule.
To quantify role-specific reliability, we compute normalized weights:
\begin{equation}
    \label{eq:role_weight}
    w_{i,k,c}^{(t)} = \frac{\exp(\beta\, s_{i,k,c}^{(t)})}{\sum_{k'} \exp(\beta\, s_{i,k',c}^{(t)})},
\end{equation}
where $\beta$ controls the sharpness of the assignment.
These weights encode the relative reliability of assigning role $k$ to agent $i$ for category $c$, and are later used as conditioning signals in the reliability-aware final reasoning stage (Section~\ref{sec:synthesis}).
This stage outputs the routing plan $\{(i,\,k_i,\,w_{i,k_i,c}^{(t)})\}_{i \in \mathcal{S}_c^{(t)}}$, which specifies, for the current query, the selected agents, their assigned roles, and their role-conditioned reliability weights.

\noindent\textbf{Role-conditioned Specialist Execution.}\label{sec:analysis}
For each query, each selected agent processes $(x^{(t)}, I^{(t)})$ independently under its designated role prompt, producing a primary answer $a_i^{(t)}$ and intermediate representations $z_i^{(t)}$ (e.g., depth maps, scene-graph structures, or reasoning traces). All outputs are written to the shared evidence pool $\mathcal{E}^{(t)}$, preserving the full evidential chain (e.g., CoT), not only final predictions, for subsequent reliability-aware final reasoning within the same step.

\noindent\textbf{Reliability-aware Final Reasoning.}\label{sec:synthesis}
For final synthesis, let $F$ denote the reasoning agent (DeepSeek-VL-R1-7B). Given the input pair $(x^{(t)}, I^{(t)})$ and the routing plan, the final prediction is produced as:
\begin{equation}
    \hat{y}^{(t)}
    =
    F\!\left(
        x^{(t)},\, I^{(t)},\,
        \mathcal{E}^{(t)},\,
        \{(a_i^{(t)}, k_i,\, w_{i,k_i,c}^{(t)})\}_{i \in \mathcal{S}^{(t)}}
      \right),
\end{equation}
where $\mathcal{E}^{(t)}$ contains all primary answers and intermediate outputs generated by the selected specialists at step $t$.
Rather than performing simple weighted voting, $F$ conducts conditional re-generation by jointly considering agreement and divergence patterns across specialist outputs. 
The intermediate representations stored in $\mathcal{E}^{(t)}$ provide structured spatial cues beyond final answers of specialists.
The weights $w_{i,k_i,c}^{(t)}$ modulate the influence of each role during final reasoning, prioritizing agents with higher estimated reliability weights for the current task category.
After producing $\hat{y}^{(t)}$, the confidence score update described in Section~\ref{sec:tto} is triggered.

\begin{figure}[t]
  \centering
  \includegraphics[width=\linewidth]{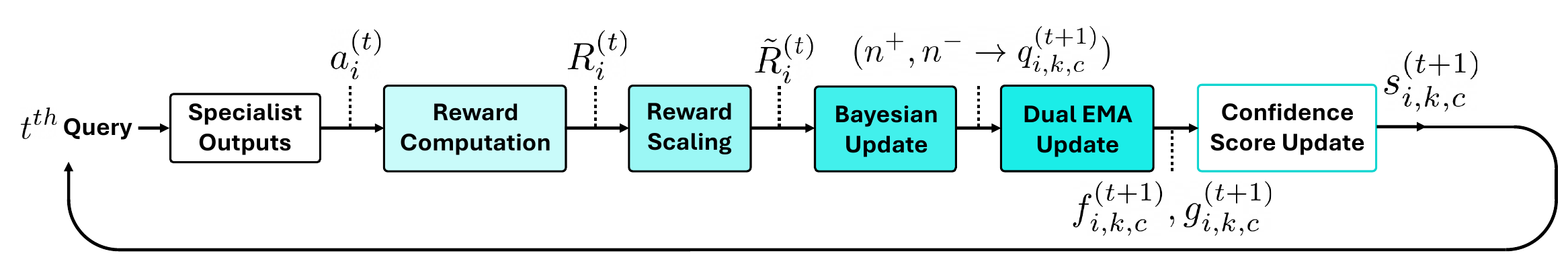}
  \caption{\textbf{Overall framework of the proposed Test-time Orchestration (TTO).}}
  \label{fig:tto}
\end{figure}
\vspace{2mm}
\subsection{Test-Time Orchestration}
\label{sec:tto}

\noindent{\textbf{Formulation.  }} For each triple $(i, k, c)$, we maintain a confidence score $s_{i,k,c}^{(t)} \in [0,1]$ quantifying the expected reliability of agent $i$ under role $k$ on task category $c$ at time $t$. 
The Test-Time Orchestration (TTO) objective is to improve the final answer accuracy over the query stream by updating these scores from observed outcomes, with zero parameter updates to any model.

\noindent\textbf{Reward Computation. }
We distinguish two supervision regimes based on whether the final reasoning output 
$\hat{y}^{(t)}$ matches the ground truth $y^{(t)}$ (agreement) or not (divergence). 
Importantly, similarity is always computed at the specialist level by comparing each agent’s output $a_i^{(t)}$ to the ground truth, regardless of the correctness of the final reasoner.

Here, $\mathrm{sim}(\cdot,\cdot)$ denotes a normalized semantic similarity between two textual answers.
In our implementation, we compute the cosine similarity between their sentence embeddings and rescale it to $[0,1]$:
\begin{equation}
    \mathrm{sim}(u,v)
    =
    \frac{1}{2}
    \left(
    \frac{\phi(u)^\top \phi(v)}
    {\|\phi(u)\|_2 \|\phi(v)\|_2}
    + 1
    \right),
    \label{eq:similarity}
\end{equation}
where $\phi(\cdot)$ denotes a sentence embedding encoder. 
For multiple-choice benchmarks, answers are first normalized to their canonical option text before computing similarity, so that semantically equivalent outputs such as ``(A) yes'' and ``yes'' are treated consistently.

A soft reward is defined as:
\begin{equation}
    R_i^{(t)} =
    \begin{cases}
        2 \cdot \mathrm{sim}(a_i^{(t)}, y^{(t)}) - 1
            & \hat{y}^{(t)} = y^{(t)}, \\[3pt]
        2 \cdot \mathrm{sim}(a_i^{(t)}, y^{(t)}) - 1 - \kappa\,\delta_i^{(t)}
            & \hat{y}^{(t)} \neq y^{(t)},
    \end{cases}
    \label{eq:reward}
\end{equation}
\vspace{-0.5mm}
where $\delta_i^{(t)}$ denotes a relative underperformance penalty applied in the divergence case. It is defined as
\begin{equation}
    \delta_i^{(t)}
    =
    \max\!\big(
    0,\,
    \mathrm{sim}(\hat{y}^{(t)}, y^{(t)})
    -
    \mathrm{sim}(a_i^{(t)}, y^{(t)})
    \big).
    \label{eq:delta}
\end{equation}

Even when the final reasoning is correct, individual specialists may still be partially incorrect; conversely, under divergence, some specialists 
may have produced answers closer to the ground truth than the final output. 
Thus, the supervision signal for each specialist is defined at the individual level, rather than being determined by the correctness of the final reasoning outcome.
In the divergence case, $\delta_i^{(t)}$ additionally penalizes agents whose outputs are less similar to the ground truth than the already incorrect final reasoning output~\cite{fung2022trust}. 
Thus, agents that underperform relative to the failed consensus incur a larger penalty, controlled by $\kappa > 0$.

To prevent premature trust divergence in sparsely observed categories, 
the reward is further scaled by a category-dependent ramp factor:
\begin{equation}
    \tilde{R}_i^{(t)} = \phi(N_c)\cdot R_i^{(t)}, \qquad
    \phi(N_c) = 1 - \exp\!\left(-\tfrac{N_c}{T}\right),
    \label{eq:reward_scaling}
\end{equation}
where $N_c$ is the cumulative query count for category $c$, and $T>0$ is a ramp temperature controlling how quickly full trust adaptation activates (smaller $T$: faster ramp-up). 
In our experiments, we set $T=5$ unless otherwise specified.

\noindent{\textbf{Bayesian Trust Update.   }}
To accumulate long-term reliability statistics for each $(i,k,c)$ triple, we maintain a Beta-Bernoulli trust model~\cite{hallyburton2024bayesian}. Here, $n^{+}_{i,k,c}(t)$ and $n^{-}_{i,k,c}(t)$ denote the cumulative soft counts of positive and negative evidence, respectively, for agent $i$ under role $k$ in category $c$ up to step $t$.

The scaled reward $\tilde{R}_i^{(t)} \in [-1,1]$ is first mapped to $[0,1]$ via $\tilde{r}_{i,k,c}^{(t)} = (\tilde{R}_i^{(t)} + 1)/2$, which can be interpreted as a fractional success probability. The sufficient statistics are then updated as:
\begin{align}
    n^{+}_{i,k,c}(t{+}1) &= n^{+}_{i,k,c}(t) + \tilde{r}_{i,k,c}^{(t)}, \\
    n^{-}_{i,k,c}(t{+}1) &= n^{-}_{i,k,c}(t) + (1 - \tilde{r}_{i,k,c}^{(t)}).
\end{align}
Intuitively, high-reward steps contribute more mass to the positive count, while low-reward steps increase the negative count, enabling gradual and noise-tolerant confidence score adaptation over time.
The posterior mean is computed as:
\begin{equation}
    q_{i,k,c}^{(t+1)} 
    =
    \frac{
        n^{+}_{i,k,c}(t+1)
    }{
        n^{+}_{i,k,c}(t+1) + n^{-}_{i,k,c}(t+1)
    }.
\end{equation}
This quantity serves as a calibrated estimate of long-run reliability.

\noindent{\textbf{Dual Exponential Moving Average.}}
To balance rapid adaptation with long-term stability, we maintain two exponential moving averages (EMA) \cite{fung2022trust} for each $(i,k,c)$ triple:
\vspace{-0.5mm}
\begin{align}
    f_{i,k,c}^{(t+1)} 
    &= 
    (1{-}\lambda_f)\, f_{i,k,c}^{(t)} 
    + 
    \lambda_f\, \tilde{R}_i^{(t)}, 
    \label{eq:ema_short}\\
    g_{i,k,c}^{(t+1)} 
    &= 
    (1{-}\lambda_g)\, g_{i,k,c}^{(t)} 
    + 
    \lambda_g\, q_{i,k,c}^{(t+1)}, 
    \label{eq:ema_long}
\end{align}
\vspace{-0.5mm}
The short-term component $f_{i,k,c}$ tracks recent reward signals $\tilde{R}_i^{(t)}$, allowing the system to quickly react to sudden performance changes. 
In contrast, the long-term component $g_{i,k,c}$ smoothly accumulates the Bayesian confidence score estimate $q_{i,k,c}$, capturing stable historical competence.

By setting $\lambda_f \gg \lambda_g$, the short-term average reacts rapidly while the long-term average evolves conservatively, realizing a dual-timescale adaptation mechanism.
The final trust score is computed as:
\begin{equation}
    s_{i,k,c}^{(t+1)} 
    = 
    \mu\, f_{i,k,c}^{(t+1)} 
    + 
    (1{-}\mu)\, g_{i,k,c}^{(t+1)} 
    + 
    \gamma\, \tilde{R}_i^{(t)}.
    \label{eq:final_score}
\end{equation}
Here, $\mu$ balances short-term responsiveness against long-term
stability, while the direct reward injection term
$\gamma\, \tilde{R}_i^{(t)}$ ensures that strong immediate feedback
can transiently influence routing before being absorbed
into the moving averages.
Updated scores are fed back into Stage-1 routing, closing the adaptation loop.

Intuitively, spatial errors are role-dependent: IVR failures can be compensated by 3D or scene-graph evidence, and vice versa. SpatiO operationalizes this by continuously re-allocating $(i,k,c)$ confidence, realizing a per-category mixture-of-experts whose short-term EMA reacts quickly to distribution shifts while the Bayesian posterior resists noisy over-reaction.

\begin{table}[t]
\centering
\caption{Full-dataset evaluation on 3DSRBench. Best results are \textbf{bold}, second best are \underline{underlined}. \textbf{Loc.}=Location, \textbf{Ori.}=Orientation, \textbf{M.O.}=Multi-object.}
\label{tab:supp_full_3dsr_app}
\scriptsize
\setlength{\tabcolsep}{2pt}
\renewcommand{\arraystretch}{1.12}
\newcolumntype{Y}{>{\centering\arraybackslash}p{0.98cm}}
\resizebox{\textwidth}{!}{%
\begin{tabular}{l*{12}{Y}|Y}
\toprule
& \multicolumn{12}{c}{3DSRBench} & \\
\cmidrule(lr){2-13}
Methods
& \makecell{Loc.\\above}
& \makecell{Height\\higher}
& \makecell{Loc.\\closer}
& \makecell{M.O.\\closer}
& \makecell{Ori.\\left}
& \makecell{M.O.\\facing}
& \makecell{Same\\dir.}
& \makecell{Ori.\\front}
& \makecell{M.O.\\VP}
& \makecell{Ori.\\VP}
& \makecell{Loc.\\next to}
& \makecell{M.O.\\parallel}
& \cellcolor{gray!20}Overall \\
\midrule
\rowcolor{gray!15}
\multicolumn{14}{l}{\textit{Without LoRA training}} \\
Llava-4D
& 51.9 & 48.7 & 58.1 & 58.9 & 50.1 & 44.2 & 50.6 & 56.1 & 20.4 & 24.8 & 57.5 & 56.3 & \cellcolor{gray!20}49.1 \\
Qwen-3.0-VL-4B
& 67.2 & 57.8 & 81.7 & 65.1 & 49.3 & 62.7 & 51.2 & 55.8 & 28.6 & 39.7 & 72.6 & 50.7 & \cellcolor{gray!20}59.2 \\
Sa2VA
& 58.2 & 48.6 & 52.2 & 59.4 & 50.1 & 58.7 & 51.2 & 54.1 & 27.1 & 28.6 & 57.5 & 51.9 & \cellcolor{gray!20}50.5 \\
SpatialRGPT
& 37.4 & 42.3 & 49.5 & 34.5 & 44.0 & 56.6 & 44.4 & 55.0 & 24.5 & 27.1 & 51.2 & 45.6 & \cellcolor{gray!20}42.7 \\
SpatialReasoner
& 69.4 & 51.3 & 73.0 & 57.1 & 49.3 & 50.3 & 47.4 & 51.2 & 29.4 & 31.8 & 69.4 & 48.3 & \cellcolor{gray!20}52.4 \\
\midrule
\rowcolor{gray!15}
\multicolumn{14}{l}{\textit{With LoRA training}} \\
Llava-4D
& 59.0 & 57.5 & 57.7 & 50.9 & 49.9 & 35.6 & 50.3 & 52.3 & 26.0 & 27.1 & 54.3 & 51.3 & \cellcolor{gray!20}49.7 \\
Qwen-3.0-VL-4B
& 62.4 & 59.1 & 79.9 & 65.7 & 50.1 & 62.4 & \underline{54.4} & 59.3 & 28.3 & 38.2 & 78.2 & 46.9 & \cellcolor{gray!20}59.1 \\
Sa2VA
& 59.4 & 43.2 & 46.5 & 48.0 & 50.1 & 62.4 & 48.0 & 48.8 & 23.6 & 26.8 & 59.3 & 61.9 & \cellcolor{gray!20}48.5 \\
SpatialRGPT
& 29.0 & 37.9 & 54.3 & 24.5 & 46.9 & 48.5 & 50.3 & 48.5 & 17.5 & 30.2 & 44.3 & 45.2 & \cellcolor{gray!20}39.8 \\
SpatialReasoner
& 69.4 & 54.3 & 71.5 & 50.3 & 53.3 & 51.7 & 45.0 & 49.4 & 30.2 & 33.6 & 68.5 & 54.0 & \cellcolor{gray!20}54.3 \\
\midrule
Gpt-5.2
& 68.6 & \textbf{68.7} & 81.3 & 74.3 & 51.0 & \underline{74.3} & \underline{63.4} & \textbf{82.6} & \underline{38.5} & \textbf{49.3} & 78.2 & 59.6 & \cellcolor{gray!20}\underline{67.2} \\
Claude opus 4.6
& 73.8 & 63.1 & 77.4 & 78.3 & 53.0 & \textbf{72.3} & 55.4 & \underline{78.3} & \textbf{42.2} & 48.2 & 60.7 & 59.0 & \cellcolor{gray!20}63.5 \\
\midrule
\rowcolor{green!15}
SpatialMAS
& \underline{73.8} & 64.2 & \underline{84.9} & \underline{74.9} & \underline{58.7} & 48.6 & 62.5 & 74.3 & 35.9 & 43.5 & \underline{79.6} & \underline{75.3} & \cellcolor{green!20}67.1 \\
\rowcolor{green!15}
\textsc{SpatiO} (Ours)
& \textbf{75.2} & \underline{65.8} & \textbf{87.3} & \textbf{76.5} & \textbf{59.4} & 49.9 & \textbf{63.7} & 76.1 & 36.8 & \underline{44.6} & \textbf{81.5} & \textbf{77.2} & \cellcolor{green!20}\textbf{72.4}\\
\bottomrule
\end{tabular}%
}
\end{table}

\section{Experiments}
In this section, we present a comprehensive evaluation of \textsc{SpatiO}, focusing on its effectiveness for adaptive agent--role orchestration in spatial reasoning.
We describe the experimental setup~(Section \ref{exp:setup}), baselines~(Section\ref{exp:main_results}), generalization~(Section\ref{exp:generalization}), and inference latency~(Section\ref{exp:latency}).

\subsection{Experimental Setup}
\label{exp:setup}

\noindent\textbf{Evaluation Data.}
We evaluate on four spatial reasoning benchmarks: STVQA-7k~\cite{batra2025spatialthinker}, CV-Bench~\cite{tong2024cambrian}, 3DSRBench~\cite{ma20253dsrbench}, and Omni3D-Bench~\cite{brazil2023omni3d}, performing inference on the full dataset for all four benchmarks.

\noindent\textbf{Baselines.}
We compare against closed-source baselines GPT-5.2~\cite{openai_gpt52_2025} and Claude-Opus~4.6~\cite{anthropic_claude_opus46_2025}, and open-source baselines Qwen3-VL-4B~\cite{bai2025qwen3}, LLaVA-4D~\cite{zhou2025llava}, SpatialReasoner~\cite{ma2025spatialreasoner}, SpatialRGPT~\cite{cheng2024spatialrgpt}, and Sa2VA~\cite{yuan2025sa2va}. All open-source baselines are evaluated both vanilla and LoRA-finetuned (rank $r{=}64$, $\alpha{=}128$) on the same 150 TTO samples, controlling for data budget and isolating orchestration-level from parameter-level adaptation.

\noindent\textbf{Ours (\textsc{SpatiO}).}
TTO adaptively estimates per-agent reliability across spatial categories using a stratified optimization set of 150 samples (30 per category, strictly separated from the evaluation set), yielding a fixed agent--role configuration for inference, with key hyperparameters $k{=}0.5$, $\mu{=}0.3$, $\gamma{=}0.3$, $\lambda_f{=}0.3$, $\lambda_g{=}0.1$, $T{=}5$, $\beta{=}5$, temperature $\tau{=}0.7$, and nucleus sampling $p{=}0.9$ for all agents (sample size ablated in Section~\ref{exp:ablation_num_samples}). We refer to this labeled calibration phase as \emph{Calibrated Trust Estimation}, distinguishing it from the per-query \emph{Test-Time Orchestration} (TTO) at inference, which requires no labels or parameter updates (full distinction in Appendix~\ref{sec:supp_terminology}).

\subsection{Main Results}
\label{exp:main_results}

We report quantitative results on STVQA-7k and CV-Bench in Tab.~\ref{tab:comparison}, and on 3DSRBench in Tab.~\ref{tab:comparison}.
Across all benchmarks, \textsc{SpatiO} consistently achieves the best overall performance, outperforming both closed-source and open-source models including LoRA-finetuned variants.
On STVQA-7k, \textsc{SpatiO} reaches 93.1 in \textit{size}, surpassing the previous best by 5.6 points.
On CV-Bench, it achieves 79.4 in \textit{count}, improving over the strongest baseline by 14 points.
On 3DSRBench, \textsc{SpatiO} obtains the best performance in 8 out of 12 subcategories.

\begin{table*}[t]

\caption{Comparison of methods on STVQA-7k (trained with 300 samples) and CV-Bench. Best results are \textbf{bold}, second best are \underline{underlined}.}
\label{tab:comparison}
\centering
\footnotesize
\setlength{\tabcolsep}{2pt}
\renewcommand{\arraystretch}{1.1}
\newcolumntype{Z}{>{\centering\arraybackslash}p{0.85cm}}
\newcolumntype{O}{>{\centering\arraybackslash}p{0.95cm}}
\resizebox{\textwidth}{!}{
\begin{tabular}{l *{9}{Z} O *{4}{Z} O}
\toprule
& \multicolumn{15}{c}{Benchmark} \\
\cmidrule(lr){2-16}
& \multicolumn{10}{c}{STVQA-7k} & \multicolumn{5}{c}{CV-Bench} \\
\cmidrule(lr){2-11} \cmidrule(lr){12-16}
Methods
& rel. & reach & size & ori. & loc. & depth & dist. & count & exist. & \cellcolor{gray!20}Overall
& count & rel. & depth & dist. & \cellcolor{gray!20}Overall \\
\midrule
\rowcolor{gray!15}
\multicolumn{16}{l}{\textit{Without LoRA training}} \\
Llava-4D
& 57.8 & 54.0 & 58.3 & 19.2 & 58.7 & 40.0 & 43.3 & 34.4 & 45.2 & \cellcolor{gray!20}52.6
& 40.8 & 61.3 & 51.3 & 54.9 & \cellcolor{gray!20}51.3 \\
Qwen-3.0-VL-4B
& 83.6 & \underline{88.0} & 70.8 & \underline{73.1} & \textbf{91.3} & \textbf{86.0} & 76.7 & 56.2 & \textbf{96.8} & \cellcolor{gray!20}82.4
& 65.0 & 87.7 & 93.8 & 83.5 & \cellcolor{gray!20}81.3 \\
Sa2VA
& 51.1 & 74.1 & 56.2 & 23.0 & 10.8 & 62.0 & 53.3 & 21.8 & 74.1 & \cellcolor{gray!20}50.0
& 59.0 & 78.0 & 80.6 & 66.1 & \cellcolor{gray!20}70.2 \\
SpatialRGPT
& 62.8 & 82.0 & 75.0 & 46.2 & 73.9 & 78.0 & 56.7 & 59.4 & 90.3 & \cellcolor{gray!20}67.1
& 51.0 & 84.4 & 55.3 & 56.5 & \cellcolor{gray!20}61.4 \\
SpatialReasoner
& 70.4 & 66.0 & 73.0 & 61.0 & 84.9 & 60.3 & \textbf{86.7} & 59.1 & 75.4 & \cellcolor{gray!20}70.6
& 58.9 & 85.6 & 83.4 & 82.3 & \cellcolor{gray!20}77.6 \\
\midrule
\rowcolor{gray!15}
\multicolumn{16}{l}{\textit{With LoRA training}} \\
Llava-4D
& 63.7 & 65.6 & 54.2 & 22.9 & 60.4 & 52.3 & 46.7 & 30.2 & 39.1 & \cellcolor{gray!20}57.2
& 61.1 & 61.6 & 82.3 & 70.8 & \cellcolor{gray!20}68.3 \\
Qwen-3.0-VL-4B
& 81.2 & 80.0 & 81.2 & 50.0 & 86.9 & 74.0 & 60.0 & 59.3 & 80.6 & \cellcolor{gray!20}77.9
& \underline{65.9} & 88.7 & \underline{94.0} & \underline{89.1} & \cellcolor{gray!20}\underline{84.4} \\
Sa2VA
& 62.3 & 76.0 & 72.9 & 53.8 & 67.4 & 74.0 & 50.0 & 68.8 & 77.4 & \cellcolor{gray!20}65.3
& 59.0 & 78.0 & 80.7 & 66.2 & \cellcolor{gray!20}70.2 \\
SpatialRGPT
& 62.1 & 84.0 & 74.0 & 37.8 & 70.9 & 83.4 & 59.3 & 66.4 & 92.0 & \cellcolor{gray!20}67.1
& 48.3 & 81.0 & 59.6 & 57.4 & \cellcolor{gray!20}61.0 \\
SpatialReasoner
& 60.2 & 78.0 & \underline{87.5} & 38.5 & 63.0 & 66.0 & 53.3 & 62.5 & 71.0 & \cellcolor{gray!20}63.4
& 61.5 & 86.9 & 82.1 & 83.1 & \cellcolor{gray!20}77.4 \\
\midrule
Gpt-5.2
& 79.7 & \textbf{92.0} & 85.4 & 69.2 & 82.6 & 82.0 & 76.6 & \underline{75.0} & \underline{96.7} & \cellcolor{gray!20}81.4
& 65.4 & \underline{95.7} & 90.4 & 87.2 & \cellcolor{gray!20}83.5 \\
Claude opus 4.6
& \underline{88.1} & 82.0 & \underline{87.5} & 69.2 & 80.4 & 80.0 & 70.0 & 71.9 & \textbf{96.8} & \cellcolor{gray!20}\underline{84.7}
& 59.2 & \textbf{94.8} & 88.8 & 86.8 & \cellcolor{gray!20}82.4 \\
\midrule
\rowcolor{green!10}
\textsc{SpatiO} (Ours)
& \textbf{88.5} & 78.0 & \textbf{93.1} & \textbf{77.5} & \underline{89.0} & \underline{85.4} & \underline{79.4} & \textbf{78.0} & 92.5 & \cellcolor{green!20}\textbf{88.2}
& \textbf{69.2} & 91.8 & \textbf{95.7} & \textbf{91.2} & \cellcolor{green!20}\textbf{86.9} \\
\bottomrule
\end{tabular}
}
\end{table*}

\begin{table}[t]
\centering
\vspace{2mm}
\caption{\textbf{Comparison on Omni3D-Bench (zero-shot).} Best: \textbf{bold}, second: \underline{underline}.}
\label{tab:omni3d}
\setlength{\tabcolsep}{8pt}
\begin{tabular}{lccc}
\toprule
\multicolumn{4}{c}{Omni3D-Bench} \\
\cmidrule(lr){1-4}
Methods & float & int & str \\
\midrule
Llava-4D & 17.2 & 3.9 & 1.2 \\
Qwen-3.0-VL-4B & \underline{23.2} & \underline{12.3} & 18.0 \\
Sa2VA & 13.6 & 2.9 & 0.0 \\
SpatialRGPT & 3.1 & 5.4 & \underline{19.2} \\
SpatialReasoner & 18.5 & 4.3 & 1.2 \\
\midrule
\rowcolor{green!10}
\textsc{SpatiO} & \textbf{40.3} & \textbf{32.9} & \textbf{20.0} \\
\bottomrule

\end{tabular}
\end{table}

\noindent\textbf{Zero-shot Performance on Omni3D-Bench.}
To assess generalization to out-of-distribution benchmarks, we evaluate on Omni3D-Bench~\cite{brazil2023omni3d} following the protocol of~\cite{marsili2025visual}. \textsc{SpatiO} is optimized on a mixed set of 150 samples (50 from each of 3DSRBench, STVQA-7k, and CV-Bench), with no Omni3D-Bench samples included. As shown in Tab.~\ref{tab:omni3d}, \textsc{SpatiO} achieves large gains across all output types, most pronounced on \texttt{float} (+17.1) and \texttt{int} (+20.6) over the strongest baseline (Qwen3-VL-4B), reflecting the effectiveness of heterogeneous orchestration for precise numerical spatial estimation and demonstrating that TTO generalizes robustly to unseen tasks and answer formats.

\vspace{-1mm}

\subsection{Generalization to Unseen Benchmarks}
\label{exp:generalization}
We further evaluate \textsc{SpatiO} on MMSI-Bench~\cite{yang2025mmsi} and MindCube~\cite{yin2025spatial}, two benchmarks not included during TTO optimization and drawn from visual domains categorically distinct from the calibration pool (\eg, autonomous driving, indoor 3D scans, egocentric video), testing whether $\mathcal{C}$ transfers beyond the calibration distribution and addressing possible benchmark-leakage concerns.
\textsc{SpatiO} achieves the highest overall accuracy on both: +19.5\%p on MMSI-Bench (largest gains on \textit{Positional Relationship}, +17.8\%, and \textit{Multi-Step Reasoning}, +29.6\%) and +1.67\%p on MindCube. Full results, category-level analysis, and the leakage discussion are in Appendix~\ref{sec:supp_generalization}.

\vspace{-1mm}

\subsection{Inference Latency}
\label{exp:latency}

\begin{table}[H]
\centering
\vspace{1mm}
\caption{Per-module inference latency (seconds per query, averaged over the full dataset from STVQA-7k). Specialists run in parallel; end-to-end latency is the maximum
across the three selected specialists plus Head and Final Reasoning Agent.}
\label{tab:single_vs_spatio}

\resizebox{\textwidth}{!}{%
\setlength{\tabcolsep}{4pt}
\begin{tabular}{l ccccc c}
\toprule
Module
& \makecell{Spatial\\Relation}
& \makecell{Distance\\Depth}
& \makecell{Size}
& \makecell{Counting}
& \makecell{Orientation}
& \cellcolor{gray!20}Avg. \\
\midrule
Head Agent (Qwen3-VL-4B)
& 2.4 & 2.6 & 2.3 & 2.5 & 2.7 & \cellcolor{gray!20}2.5 \\
\midrule
Role 1 · Implicit Visual
& 14.2 & 15.8 & 12.1 & 17.3 & 20.6 & \cellcolor{gray!20}16.0 \\
Role 2 · Explicit 3D
& 2.8 & 8.4 & 2.1 & 2.3 & 7.9 & \cellcolor{gray!20}4.5 \\
Role 3 · Scene Graph
& 9.3 & 5.2 & 4.8 & 8.6 & 4.1 & \cellcolor{gray!20}6.4 \\
\midrule
Final Reasoning Agent (DSR-7B)
& 4.8 & 5.1 & 3.9 & 4.2 & 4.5 & \cellcolor{gray!20}4.5 \\
\midrule
\rowcolor{green!10}
\textbf{End-to-end (\textsc{SpatiO})}
& 31.5 & 31.9 & 18.3 & 26.0 & 32.7 & \cellcolor{green!20}28.1 \\
\bottomrule
\end{tabular}
}
\end{table}

The heterogeneous multi-agent architecture of \textsc{SpatiO} introduces additional inference cost relative to single-model baselines. Tab.~\ref{tab:single_vs_spatio} directly compares \textsc{SpatiO} against each single-model baseline on accuracy, average per-query latency, throughput, and peak VRAM on 3DSRBench (single A100 80GB GPU).\\
\textsc{SpatiO} improves accuracy over the strongest single-model baseline, SpatialReasoner, from 65.6\% to 84.7\% (+19.1\%p) while \emph{reducing} average latency from 25.8\,s to 19.8\,s. This is possible because the three specialist roles execute \emph{in parallel} and the Head Agent dynamically selects the most reliable specialist combination per query category, invoking heavyweight specialists like SpatialReasoner only when the task demands it, rather than for every query; throughput and VRAM are correspondingly higher than any individual specialist alone, reflecting the cost of running multiple specialists in parallel.
Per-module and per-category latency breakdowns are provided in Appendix~\ref{sec:single_vs_spatio}.

\section{Analysis}

We further analyze the behavior of TTO along three axes: (1) the contribution of each optimization component and the effect of calibration sample size, (2) how confidence scores evolve over time, and (3) whether trust scores transfer across structurally different benchmarks. Due to space constraints, full results and discussion for all three axes are provided in Appendix~\ref{sec:supp_tto_analysis}; we summarize the key findings below.

\noindent\textbf{TTO component ablation.}
Progressively adding (1)~reward computation, (2)~reward scaling, (3)~Bayesian trust update, and (4)~dual EMA each consistently improves accuracy (Appendix Tab.~\ref{tab:ablation_tto_step}), with the Bayesian update yielding the largest single gain by grounding agent selection in accumulated evidence. A complementary leave-one-out ablation (Appendix Tab.~\ref{tab:component_ablation}) confirms each component is individually necessary, with the $\delta_i$ underperformance penalty and the Final Reasoning Agent contributing the largest individual drops when removed.
Varying the TTO calibration sample size shows accuracy peaks at 150 samples (91.30\%) before slightly degrading, motivating our default choice of 150 (30 per category; Appendix Fig.~\ref{fig:ablation_num_samples}). We additionally verify robustness to the Top-$k$, $\beta$, and role-assignment hyperparameters, and to substituting the agent pool or role prompts (Appendix Tabs.~\ref{tab:hyperparam_sensitivity},~\ref{tab:pool_role_variants}).

\noindent\textbf{Confidence score evolution.}
Tracking per-agent confidence trajectories within a fixed category (Appendix Fig.~\ref{fig:confidence_evolution}) shows that TTO progressively differentiates reliable agents from weaker ones within each role--category pair, realizing a per-category mixture-of-experts at test time without any parameter update.

\noindent\textbf{Cross-benchmark generalization.}
Calibrating on 3DSRBench (3D-dominant, fine-grained categories) and evaluating zero-shot on STVQA-7k outperforms the reverse direction, calibrating on CV-Bench (2D-dominant), by +7.1\% (Appendix Tab.~\ref{tab:supp_cross_cv},~\ref{tab:supp_cross_3dsr}).
This asymmetry suggests that trust profiles calibrated on harder, finer-grained 3D categories transfer more readily to simpler 2D-dominant tasks than the reverse.

\vspace{-1mm}
\section{Conclusion}
In this work, we introduced \textsc{SpatiO}, a heterogeneous multi-agent framework for spatial reasoning that coordinates vision-language specialists with complementary inductive biases. To enable effective collaboration among agents, we proposed Test-Time Orchestration (TTO), a lightweight optimization mechanism that dynamically calibrates agent reliability during inference based on accumulated interaction evidence.
Extensive experiments across diverse spatial reasoning benchmarks demonstrate that \textsc{SpatiO} consistently improves performance over both closed-source and open-source baselines, suggesting that dynamically orchestrating heterogeneous reasoning agents at test time is a promising direction for spatial reasoning in vision-language models without additional training.

\section*{Acknowledgements}

This work was supported by the National Research Foundation of Korea (NRF) (RS-2026-25488668, 20\%), the Institute of Information \& Communications Technology Planning \& Evaluation (IITP) under the Leading Generative AI Human Resources Development grant (IITP-2026-RS-2024-00397085, 20\%), the Artificial Intelligence Star Fellowship Support Program to Nurture the Best Talents grant (IITP-2026-RS-2025-02304828, 40\%), and the IITP-ICT Creative Consilience Program
grant (IITP-2026-RS-2020-II201819, 20\%), funded by the Korea government (MSIT).

\clearpage
\bibliographystyle{splncs04}
\bibliography{main}

\clearpage
\setcounter{table}{0}
\renewcommand{\thetable}{A\arabic{table}}
\setcounter{figure}{0}
\renewcommand{\thefigure}{A\arabic{figure}}
\appendix

\begin{center}
  {\LARGE\bfseries Supplementary Material}\\[6pt]
  {\large\bfseries \textsc{SpatiO}: Adaptive Test-Time Orchestration of\\
  Vision-Language Agents for Spatial Reasoning}
\end{center}

\vspace{3mm}
\hrule
\vspace{5mm}

\section*{Appendix Overview}

\noindent
In Section~A (p.~\pageref{sec:supp_exp_details}), we present additional experimental details, hyperparameter settings, and a clarification of TTO terminology.
In Section~B (p.~\pageref{sec:supp_sampling}), we provide details on the optimization data sampling strategy.
In Section~C (p.~\pageref{sec:supp_experiments}), we report additional quantitative results, including TTO component ablation (cumulative and leave-one-out), hyperparameter sensitivity, agent pool/role design validation, confidence score evolution, cross-benchmark generalization, generalization to MMSI-Bench and MindCube with a benchmark-leakage discussion, comparison with prior multi-agent frameworks, inference latency (including a direct single-model comparison), full-dataset evaluation, and Head Agent ablation.
In Section~D (p.~\pageref{sec:supp_limitations_futureworks}), we discuss limitations and future directions.
In Section~E (p.~\pageref{sec:supp_qualitative}), we present qualitative results.
In Section~F (p.~\pageref{sec:supp_prompt_rationale}), we provide the prompt design rationale and full system prompts for each agent.

\vspace{5mm}

\section{Experimental Details}
\label{sec:supp_exp_details}

\subsection{Specialist Agents for Spatial Reasoning}
\vspace{1mm}
\begin{itemize}
  \item \textbf{Qwen3-VL-4B}~\cite{bai2025qwen3}: a general-purpose vision-language model that delivers advanced multimodal reasoning and visual understanding through tightly integrated joint text-vision processing, strong 2D/3D spatial grounding, and extended contextual comprehension. This model serves as the head agent for adaptive role assignment within the framework, where it directs incoming spatial queries to the relevant specialist models according to the query type.
  \item \textbf{LLaVA-4D}~\cite{zhou2025llava}: the first vision-language model to understand scenes in four dimensions, encompassing both three-dimensional spatial positions and a temporal axis. This enables the model to differentiate dynamic objects from static backgrounds and to reason about motion across time.
  \vspace{0.5mm}
  \item \textbf{SpatialReasoner}~\cite{ma2025spatialreasoner}: employs explicit three-dimensional representations, including locations and orientations, as a structured interface for multi-step spatial reasoning from two-dimensional images. The model undergoes a two-stage training process: supervised fine-tuning to establish three-dimensional perception, followed by reinforcement learning to develop generalizable spatial reasoning capabilities.
  \vspace{0.5mm}
  \item \textbf{SpatialRGPT}~\cite{cheng2024spatialrgpt}: incorporates a depth-aware plugin and a three-dimensional scene graph-based data pipeline to support region-level spatial reasoning. This approach allows the model to infer metric relationships, including distance and relative depth, which exceed the capabilities of appearance-based cues alone.
  \vspace{0.5mm}
  \item \textbf{Sa2VA}~\cite{yuan2025sa2va}: combines SAM2-based segmentation with language understanding in a unified model, enabling pixel-level grounding of spatial references expressed in text. This integrated architecture enables precise localization of objects described in natural language and reduces reliance on bounding-box approximations.
\end{itemize}

\subsection{Reasoner Agent}

\noindent\textbf{DeepSeek-R1-7B}~\cite{guo2025deepseek}.
The Reasoner Agent, referred to as DeepSeek-VL-R1-7B in the main paper, corresponds to \texttt{DeepSeek-R1-Distill-Qwen-7B}. Built on a Qwen-7B backbone, it processes both visual and textual inputs. As a distilled variant of DeepSeek-R1, it inherits strong chain-of-thought reasoning cultivated through reinforcement learning, making it well-suited for aggregating specialist outputs, resolving evidential conflicts via TTO reliability weights, and producing the final answer.

\subsection{Spatial Reasoning Benchmarks}
\vspace{2mm}

\noindent{\textbf{STVQA} \cite{batra2025spatialthinker}} is a spatial visual question answering benchmark introduced with SpatialThinker method.
The dataset is constructed from scene graphs derived from Visual Genome \cite{krishna2017visual} and contains approximately 7.5K multiple-choice VQA pairs designed to evaluate both 2D and 3D spatial reasoning abilities.
It covers diverse spatial reasoning types including relations, size, orientation, distance, depth, reach, location, count, and existence.

\vspace{1mm}
\noindent{\textbf{CV-Bench} \cite{tong2024cambrian}} is a vision-centric evaluation benchmark to evaluate the spatial reasoning capabilities of multimodal models. 
The benchmark is constructed by repurposing standard vision datasets, including ADE20K \cite{zhou2019semantic}, COCO \cite{lin2014microsoft}, and Omni3D \cite{brazil2023omni3d}, leveraging their rich ground-truth annotations. 
Based on these annotations, natural language questions are formulated to probe the fundamental visual and spatial understanding of the models. 
CV-Bench evaluates both 2D and 3D spatial reasoning abilities: 2D understanding is assessed through tasks such as spatial relationship recognition and object counting, while 3D understanding is evaluated via depth ordering and relative distance estimation. 

\vspace{1mm}
\noindent{\textbf{3DSRBench} \cite{ma20253dsrbench}} is a comprehensive benchmark designed to evaluate the 3D spatial reasoning capabilities of large multimodal models.
It contains 2,700+ manually annotated visual question–answer pairs spanning multiple reasoning categories such as height comparison, object orientation, relative location, and multi-object spatial relationships.
The benchmark includes both real-world images and synthetic scenes rendered from different camera viewpoints, enabling evaluation of robustness under varying perspectives.

\vspace{1mm}
\noindent\textbf{MMSI-Bench}~\cite{yang2025mmsi} is a benchmark designed to evaluate multi-step spatial intelligence in multimodal models. It comprises 1,000 visual question--answer pairs spanning four reasoning categories: multi-step reasoning, positional relationships, motion, and attribute recognition. Unlike benchmarks that probe isolated spatial predicates, MMSI-Bench emphasizes compositional reasoning chains that require integrating multiple spatial inferences in sequence.

\vspace{1mm}
\noindent\textbf{Omni3D-Bench}~\cite{brazil2023omni3d}: Repurposed as an open-ended spatial estimation protocol by~\cite{marsili2025visual}, requiring metric outputs in three formats (\texttt{float}, \texttt{int}, \texttt{str}).

\subsection{Implementation Details}
Our method introduces several hyperparameters, including \(k\), \(\mu\), \(\gamma\), \(\lambda_f\), \(\lambda_g\), \(T\), and \(\beta\).
Unless otherwise specified, we set \(k = 0.5\), \(\mu = 0.3\), \(\gamma = 0.3\), \(\lambda_f = 0.3\), \(\lambda_g = 0.1\), \(T = 5 \) and \( \beta = 5 \).
For generation, all the Head Agent, Specialist Agents and the Reasoner Agent employ stochastic decoding with temperature \(0.7\) and nucleus sampling (\(p=0.9\)).
The maximum number of generated tokens is set to \(64\) for the Head Agents and \(1024\) for the Specialist Agents and Reasoner Agent.

\vspace{2mm}
\subsection{Clarifying TTO Terminology: Calibration vs.\ Test-Time Orchestration}
\label{sec:supp_terminology}

We clarify a terminological distinction raised during review. \textsc{SpatiO} involves two logically separate phases that we both previously referred to under the umbrella of ``test-time optimization,'' which overstates the analogy to conventional, unsupervised test-time adaptation (TTA):

\begin{itemize}[leftmargin=1em, itemsep=2pt]
  \item \textbf{Calibrated Trust Estimation.} A one-time phase that uses a small, strictly held-out, \emph{labeled} set (150 samples, 30 per category) to initialize the per-(agent, role, category) confidence scores $s_{i,k,c}$ via Eqs.~\ref{eq:reward}--\ref{eq:final_score}. This is closer to a parameter-free few-shot calibration of orchestration weights than to unsupervised TTA (\eg, TENT~\cite{liang2025comprehensive}, MEMO~\cite{sun2020test}), which adapt on \emph{unlabeled} test inputs.
  \item \textbf{Test-Time Orchestration (TTO).} The per-query construction procedure that actually runs at inference: given the calibrated confidence scores, the Head Agent selects the Top-3 agents, assigns roles, and the Final Reasoning Agent integrates evidence (Sec.~\ref{sec:tto}). This step requires no labels and no parameter updates, and is what we mean by ``test-time'' throughout the paper.
\end{itemize}

\noindent In other words, \emph{calibration} is what is supervised; \emph{orchestration} is what happens at test time. We retain the name ``TTO'' for the overall framework since the confidence scores continue to be updated online from the query stream after calibration (Sec.~\ref{sec:tto}), but we recommend this distinction be kept in mind when comparing against unsupervised TTA baselines. The calibrated setting is most directly applicable to domains such as robotics or 3D scene understanding, where a small labeled set can be collected once from the target environment or sensor configuration and reused across deployment.

\vspace{4mm}

\section{Optimization Data Sampling}
\label{sec:supp_sampling}

\noindent\textbf{TTO Optimization Set.}
For each benchmark, 150 samples are used for TTO, with strict separation from the evaluation set. Sampling is stratified over the five categories of $\mathcal{C}$ (30 instances per category), ensuring every category receives sufficient signal for the confidence scores $s_{i,k,c}^{(t)}$ to converge.
For Omni3D-Bench (zero-shot), the optimization set consists of 50 samples each from 3DSRBench, STVQA-7k, and CV-Bench (150 total); no Omni3D-Bench samples are included, consistent with~\cite{marsili2025visual}.

\vspace{1mm}
\noindent\textbf{LoRA Fine-tuning Baseline.}
All LoRA baselines are fine-tuned on the same 150 samples, with rank $r{=}64$ and $\alpha{=}128$, controlling for data budget.

\vspace{1mm}
\noindent\textbf{Full-dataset Evaluation.}
For full-dataset experiments, the same 150-sample optimization set is used; the evaluation set remains strictly disjoint.

\vspace{2mm}

\section{Additional Quantitative Results}
\label{sec:supp_experiments}

\subsection{TTO Ablation, Confidence Evolution, and Cross-Benchmark Generalization}
\label{sec:supp_tto_analysis}

\begin{figure}[t]
\centering

\begin{subfigure}[t]{0.33\linewidth}
    \centering
    \includegraphics[width=1.1\linewidth]{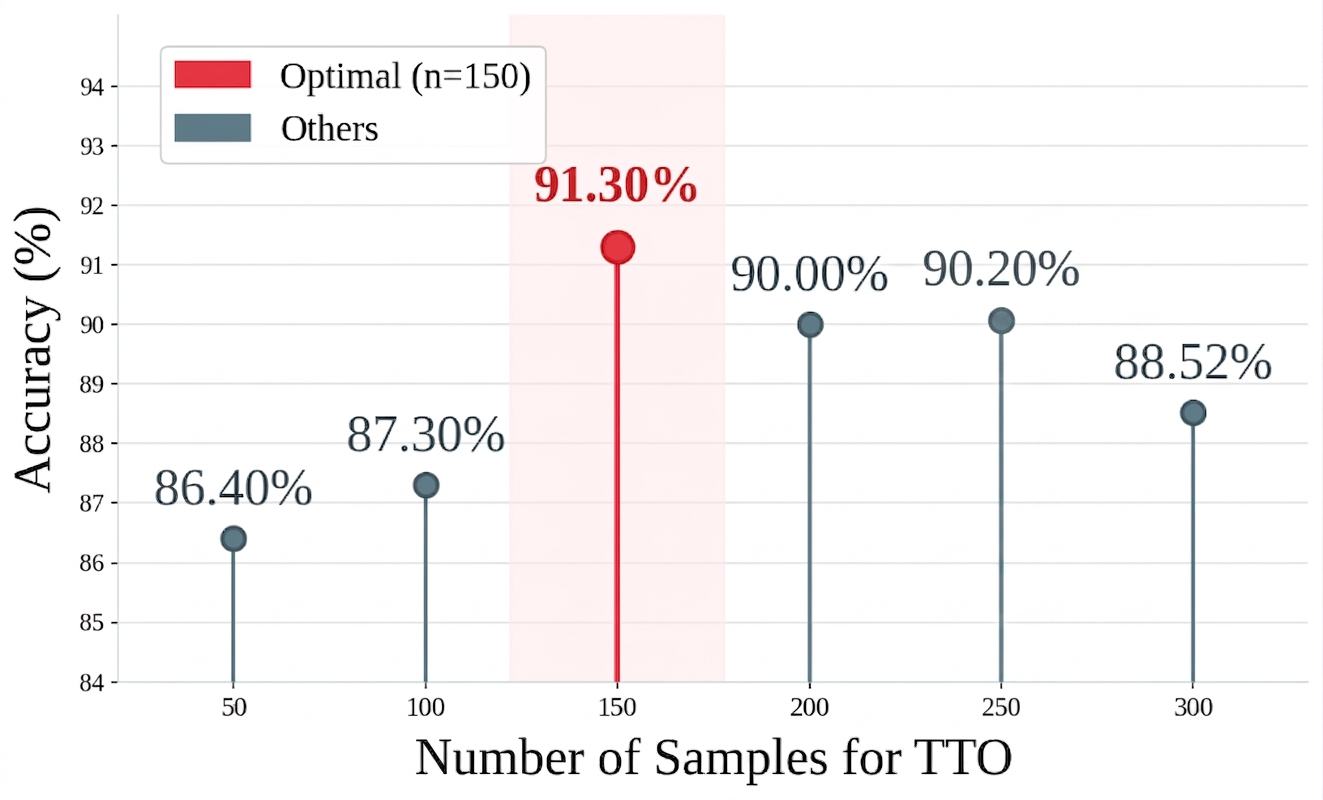}
    \caption{Overall Performance on CV-Bench according to TTO sample size.}
    \label{fig:ablation_num_samples}
\end{subfigure}
\hfill
\begin{subfigure}[t]{0.65\linewidth}
    \centering
    \includegraphics[width=\linewidth]{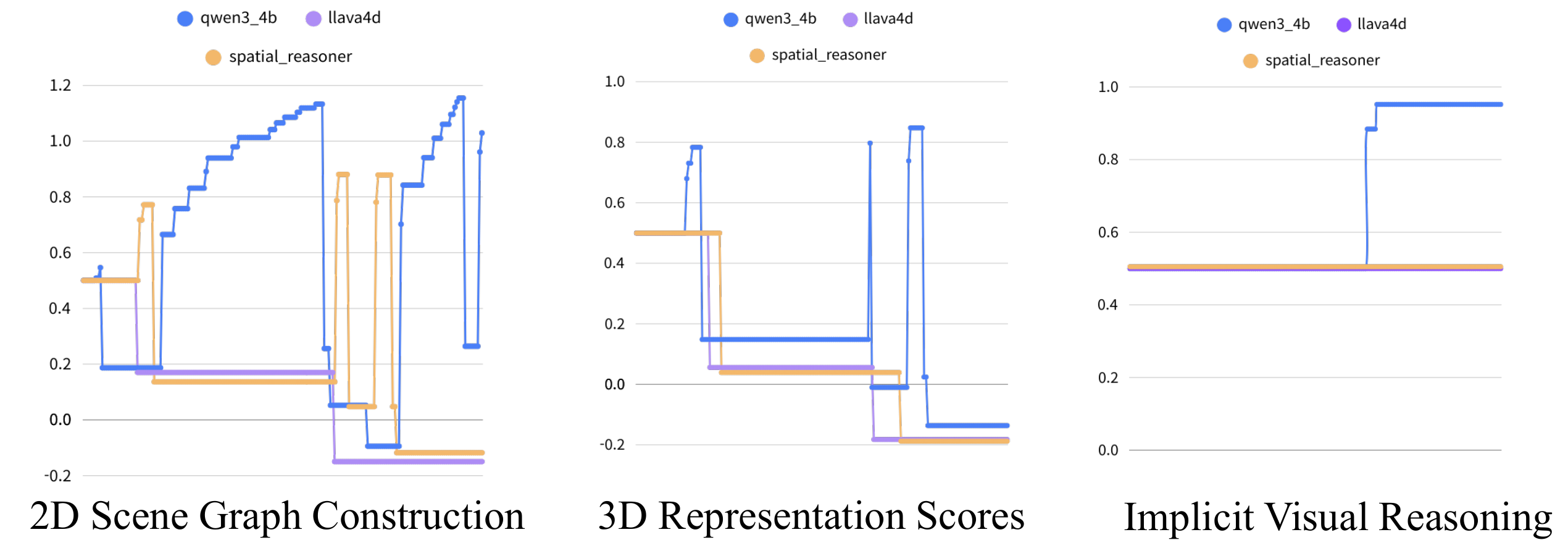}
    \caption{Evolution of agent confidence scores during TTO optimization.}
    \label{fig:confidence_evolution}
\end{subfigure}

\caption{\textbf{Analysis of Test-Time Orchestration (TTO).} 
\textit{Left}: Effect of the number of samples used in TTO optimization.
\textit{Right}: Evolution of agent confidence scores for the \textit{counting} category. Each curve represents the confidence score of an agent assigned to one of three roles for a fixed task category}
\label{fig:tto_analysis}
\end{figure}

\noindent\textbf{Effect of TTO optimization sample size.}
\label{exp:ablation_num_samples}
We analyze how the number of samples used during TTO optimization affects performance.
We randomly sample 500 instances from CV-Bench for evaluation and vary the optimization sample size from 50 to 300.
As shown in Fig.~\ref{fig:ablation_num_samples}, performance improves as sample size increases up to 150, reaching the best accuracy of 91.30\%.
Beyond 150 samples, performance slightly degrades, suggesting diminishing returns once sufficient signal is obtained for each category in the stratified optimization set.
We therefore adopt 150 samples (30 per category) as the default TTO optimization size throughout all experiments.

\vspace{2mm}
\noindent\textbf{Contribution of each TTO component.}
\label{exp:ablation_tto}
We analyze the contribution of each component in the TTO pipeline by progressively adding:
(1)~reward computation, (2)~reward scaling, (3)~Bayesian trust update, and (4)~dual exponential moving average.
Tab.~\ref{tab:ablation_tto_step} reports results on 500 sampled CV-Bench instances with 150 optimization samples.
Each component consistently improves performance.
Reward scaling provides a stable optimization signal by preventing premature trust divergence in sparsely observed categories.
The Bayesian confidence score update yields the largest single gain, enabling reliability-aware agent selection grounded in accumulated evidence.
The dual EMA further stabilizes confidence estimation by balancing short-term responsiveness against long-term reliability.

\vspace{2mm}

\begin{table}[h]
\centering
\caption{\textbf{Ablation on each step used in TTO optimization.} Results on 500 sampled CV-Bench instances with 150 optimization samples.}
\label{tab:ablation_tto_step}
\footnotesize
\setlength{\tabcolsep}{8pt}
\begin{tabular}{lc}
\toprule
\multicolumn{2}{c}{CV-Bench} \\
\midrule
\textbf{Methods} & \textbf{Accuracy} \\
\midrule
\textsc{SpatiO} & 86.10 \\
\textsc{SpatiO} + (step1) & 73.02 \\
\textsc{SpatiO} + (step2) & 79.66 \\
\textsc{SpatiO} + (step3) & 88.12 \\
\textsc{SpatiO} + (step4) & \textbf{89.86} \\
\bottomrule
\end{tabular}
\end{table}

\vspace{2mm}
\noindent\textbf{Component-wise (leave-one-out) ablation.}
While the table above adds components \emph{cumulatively}, we additionally isolate the contribution of each component by \emph{removing or replacing} it from the full model individually (Tab.~\ref{tab:component_ablation}).
Removing the $\delta_i$ underperformance penalty (e) causes the largest single drop (87.4 $\to$ 72.0), confirming that penalizing specialists who underperform relative to a failed consensus is critical for sharp trust differentiation.
Replacing the Final Reasoning Agent with simple majority voting (d) also substantially degrades performance (77.2), showing that conditional re-generation over the evidence pool, rather than vote counting, is responsible for much of the gain.
Uniform weights (a), equivalent to disabling TTO entirely, perform only marginally better than the homogeneous baseline in Fig.~\ref{fig:per_task_radar}(b), underscoring that reliability-aware orchestration is the primary source of improvement.

\begin{table}[h]
\centering
\caption{\textbf{Component-wise ablation on CV-Bench.} Each row removes/replaces a single TTO component from the full model, isolating its individual contribution (cf.\ Tab.~\ref{tab:ablation_tto_step}, which adds components cumulatively).}
\label{tab:component_ablation}
\footnotesize
\setlength{\tabcolsep}{8pt}
\begin{tabular}{lc}
\toprule
\textbf{Variant} & \textbf{Acc.\ (\%)} \\
\midrule
\rowcolor{green!10}
\textsc{SpatiO} (full)                  & \textbf{87.4} \\
(a) Uniform weights (no TTO)            & 75.9 \\
(b) w/o short-term EMA                  & 83.2 \\
(c) w/o long-term EMA                   & 81.5 \\
(d) Majority voting (no Reasoner)       & 77.2 \\
(e) w/o $\delta_i$ penalty              & 72.0 \\
\bottomrule
\vspace{1mm}
\end{tabular}
\end{table}

\vspace{2mm}
\noindent\textbf{Confidence score evolution during TTO.}
To illustrate how TTO adapts agent selection over time, Fig.~\ref{fig:tto_analysis}(b) visualizes confidence score trajectories for multiple agents within the \textit{counting} category of CV-Bench.
Each agent is initialized at 0.5 and updated over 200 steps via the reward-driven confidence mechanism; steps outside \textit{counting} appear as flat segments since updates are category-gated.
As optimization progresses, update magnitudes grow due to the reward scaling mechanism~(Eq.~\ref{eq:reward_scaling}), which amplifies as cumulative category queries accumulate.
Role-wise dynamics differ: for \textit{Scene Graph Construction} (Role~3), Qwen3-VL-4B consistently produces more reliable outputs and its confidence rises while others decay; for \textit{Implicit Visual Reasoning} (Role~1), Qwen3-VL-4B dominates from early steps.
These trajectories demonstrate how TTO progressively differentiates reliable agents from weaker ones within each role--category pair, realizing a per-category mixture-of-experts at test time without any parameter update.

\vspace{1mm}
\noindent\textbf{Cross-benchmark generalization.}
\label{sec:supp_cross}
A key question is whether the trust scores learned during TTO optimization on one benchmark transfer to a structurally different target benchmark. We evaluate two cross-benchmark settings:

\begin{itemize}[leftmargin=1em, itemsep=1pt]
  \item \textbf{CV-Bench $\to$ STVQA-7k}: optimize on CV-Bench (2D-dominant,
    multiple-choice), evaluate zero-shot on STVQA-7k.
  \item \textbf{3DSRBench $\to$ STVQA-7k}: optimize on STVQA (mixed 2D/3D), evaluate
    zero-shot on 3DSRBench (3D-dominant, fine-grained subcategories).
\end{itemize}

\begin{table}[h]
\centering
\caption{Zero-shot generalization on STVQA-7k optimized on CV-Bench (150 samples).
Best: \textbf{bold}, second: \underline{underlined}.}
\label{tab:supp_cross_cv}
\resizebox{\textwidth}{!}{%
\setlength{\tabcolsep}{4pt}
\begin{tabular}{l ccccccccc c}
\toprule
Method
& rel. & reach & size & ori. & inst.loc. & depth & dist. & count & exist.
& \cellcolor{gray!20}Overall \\
\midrule
Llava-4D        & 56.9 & 66.0 & 45.8 & 15.3 & 50.0 & 72.0 & 40.0 & 40.6 & 90.3 & \cellcolor{gray!20}55.9 \\
Qwen-3.0-VL-4B  & \underline{75.7} & \underline{82.0} & 77.5 & \textbf{76.9} & \textbf{80.4} & \textbf{85.0} & \textbf{83.3} & \textbf{68.7} & \underline{93.5} & \cellcolor{gray!20}\underline{78.7} \\
Sa2VA           & 72.0 & \underline{82.0} & \underline{79.1} & 69.2 & \textbf{80.4} & 78.0 & 56.6 & 62.5 & \textbf{96.7} & \cellcolor{gray!20}74.1 \\
SpatialRGPT     & 34.2 & 36.0 & 31.2 & 19.2 & 23.9 & 24.0 & 33.3 &  46.2 & 51.6 & \cellcolor{gray!20}33.3 \\
SpatialReasoner & 72.0 & \underline{82.0} & \underline{79.1} & 65.3 & 73.9 & 72.0 & 63.3 & 46.8 & 74.1 & \cellcolor{gray!20}71.6 \\
\midrule
\rowcolor{green!10}
\textsc{SpatiO} (Ours) & \textbf{79.3} & \textbf{85.5} & \textbf{85.0} & \underline{73.1} & \underline{76.1} & \underline{84.0} & \underline{73.3} & \underline{63.1} & 87.1 & \cellcolor{green!20}\textbf{79.3} \\
\bottomrule
\end{tabular}
}
\end{table}
\begin{table}[H]
\centering
\caption{Zero-shot generalization on STVQA-7k optimized on 3DSRBench (150 samples).
Best: \textbf{bold}, second: \underline{underlined}.}
\label{tab:supp_cross_3dsr}
\resizebox{\textwidth}{!}{%
\setlength{\tabcolsep}{4pt}
\begin{tabular}{l ccccccccc c}
\toprule
Method
& rel. & reach & size & ori. & inst.loc. & depth & dist. & count & exist.
& \cellcolor{gray!20}Overall \\
\midrule
Llava-4D        & 60.6 & 66.0 & 54.1 & 15.3 & 47.8 & 26.0 & 43.3 & 37.5 & 54.8 & \cellcolor{gray!20}53.4 \\
Qwen-3.0-VL-4B  & \underline{81.7} & \underline{82.0} & \underline{85.4} & \textbf{76.9} & \underline{78.2} & \underline{80.0} & \underline{73.3} & \underline{62.5} & \textbf{96.7} & \cellcolor{gray!20}\underline{80.9} \\
Sa2VA           & 68.0 & \underline{82.0} & 75.0 & 57.6 & 73.9 & 60.0 & 56.6 & \textbf{65.6} & \textbf{96.7} & \cellcolor{gray!20}69.6 \\
SpatialRGPT     & 70.0 & 60.4 & 69.2 & 35.0 & 43.1 & 34.4 & 35.8 & 56.3 & 88.0 & \cellcolor{gray!20}62.2 \\
SpatialReasoner & 71.0 & 80.0 & 77.1 & 53.8 & 73.9 & 54.0 & 60.0 & 43.8 & 77.4 & \cellcolor{gray!20}68.9 \\
\midrule
\rowcolor{green!10}
\textsc{SpatiO} (Ours) & \textbf{86.4} & \textbf{88.4} & \textbf{88.9} & \underline{76.6} & \textbf{89.0} & \textbf{83.4} & \textbf{79.5} & 55.4 & \underline{96.0} & \cellcolor{green!20}\textbf{86.4} \\
\bottomrule
\end{tabular}
}
\end{table}

These results reveal an asymmetric generalization pattern across the two
optimization sources. When optimizing on CV-Bench (2D-dominant, multiple-choice), \textsc{SpatiO} achieves 79.3\% on STVQA-7k, a slight drop compared to in-domain performance. By contrast, optimizing on 3DSRBench (3D-dominant, fine-grained spatial categories) yields 86.4\% on STVQA-7k, surpassing the CV-Bench-optimized setting by +7.1\%.

We attribute this asymmetry to the richer spatial complexity of 3DSRBench: its fine-grained 3D reasoning requirements force the Bayesian trust estimator to build a more discriminative per-category reliability profile across a wider range of spatial predicates. Since the per-category confidence scores $s_{i,k,c}^{(t)}$ operate over the unified taxonomy $\mathcal{C}$ shared across benchmarks, an agent reliability profile calibrated on harder 3D categories transfers directly to simpler 2D-dominant tasks, but not necessarily the reverse. CV-Bench's shallower spatial diversity provides insufficient signal to calibrate agent weights for the more demanding depth, orientation, and multi-object reasoning categories present in STVQA-7k, whereas 3DSRBench-optimized weights already encode reliable estimates for these harder categories and generalize accordingly.

\vspace{2mm}
\noindent\textbf{Hyperparameter sensitivity.}
Tab.~\ref{tab:hyperparam_sensitivity} sweeps the Top-$k$ specialist count, the role-assignment sharpness $\beta$, and the role-assignment strategy on 3DSRBench. We select $k{=}3$ for a favorable accuracy--efficiency trade-off, despite $k{=}4$ yielding a marginally higher accuracy (75.1 vs.\ 73.3): each additional active specialist adds a parallel inference branch and a corresponding latency/VRAM cost (Sec.~\ref{tab:single_vs_spatio}). For role assignment, trust-based assignment (our default) outperforms both a fixed role mapping and a random one, confirming that letting confidence scores determine which agent occupies which role is itself a source of gain, independent of the trust-weighted final reasoning stage.

\begin{table}[H]
\centering
\caption{\textbf{Hyperparameter sensitivity on 3DSRBench (\%).} Gray = default (ours). $k{=}4$ yields marginally higher accuracy than $k{=}3$, but we adopt $k{=}3$ for a better accuracy--efficiency trade-off (Sec.~\ref{exp:setup}).}
\label{tab:hyperparam_sensitivity}
\setlength{\tabcolsep}{5pt}
\begin{tabular}{ccccc}
\multicolumn{5}{c}{\textbf{Top-}$k$} \\
\toprule
$k{=}1$ & $k{=}2$ & \cellcolor{gray!20}$k{=}3$ & $k{=}4$ & $k{=}5$ \\
\midrule
47.8 & 62.0 & \cellcolor{gray!20}73.3 & 75.1 & 69.7 \\
\bottomrule
\end{tabular}
\hfill
\begin{tabular}{ccccc}
\multicolumn{5}{c}{$\boldsymbol{\beta}$} \\
\toprule
0.3 & 0.5 & \cellcolor{gray!20}0.7 & 0.9 & 1.0 \\
\midrule
71.3 & 68.7 & \cellcolor{gray!20}73.3 & 73.1 & 71.4 \\
\bottomrule
\end{tabular}
\hfill
\begin{tabular}{ccc}
\multicolumn{3}{c}{\textbf{Role Assignment}} \\
\toprule
Fixed & Rand. & \cellcolor{gray!20}Trust \\
\midrule
64.8 & 70.2 & \cellcolor{gray!20}73.3 \\
\bottomrule
\end{tabular}
\end{table}

\vspace{1mm}
\noindent\textbf{Sensitivity to agent pool and role design.}
A natural concern is whether \textsc{SpatiO}'s gains depend on the specific five-agent pool or the three hand-designed roles. Tab.~\ref{tab:pool_role_variants} shows that substituting our human-designed role prompts with LLM-generated ones (using an off-the-shelf LLM to draft the role instructions from a short natural-language description) incurs only a small drop (73.3 $\to$ 71.4 on 3DSRBench), indicating that the method is not overly sensitive to manual prompt engineering. Likewise, replacing two of the five agents with alternative backbones (Qwen2-VL, InternVL2; averaged over 3 runs) also causes only a modest decrease (70.9), suggesting that \emph{architectural complementarity}, rather than the identity of any particular agent, is the key criterion for pool construction (cf.\ Appendix~\ref{sec:supp_limitations_futureworks}).

\begin{table}[h]
\centering
\caption{\textbf{Agent pool and role design variants.} Substituting human-designed roles with LLM-generated roles, or swapping 2 of the 5 agents for alternatives (Qwen2-VL, InternVL2; averaged over 3 runs), causes only a small drop, confirming that architectural complementarity, not the specific pool/role choice, drives the gains.}
\label{tab:pool_role_variants}
\footnotesize
\setlength{\tabcolsep}{8pt}
\begin{tabular}{lcc}
\toprule
\textbf{Variant} & \textbf{3DSRBench} & \textbf{CV-Bench} \\
\midrule
\rowcolor{green!10}
Original pool + human roles                    & \textbf{73.3} & \textbf{87.4} \\
Replace w/ LLM-generated roles                  & 71.4 & 85.9 \\
$-$2 rand.\ agents $+$2 new agents (3 runs)     & 70.9 & 85.7 \\
\bottomrule
\end{tabular}
\end{table}

\vspace{1mm}
\subsection{Generalization to MMSI-Bench and MindCube}
\label{sec:supp_generalization}

\noindent\textbf{Benchmark leakage clarification.}
Because the 150-sample calibration set is drawn from 3DSRBench, STVQA-7k, and CV-Bench, a natural concern is that any ``zero-shot'' gains on benchmarks sharing visual sources with the calibration pool may not reflect genuine generalization. We address this in two ways. First, Omni3D-Bench (Sec.~\ref{exp:main_results}) evaluates a fundamentally different \emph{task family} (open-ended metric 3D estimation with \texttt{float}/\texttt{int}/\texttt{str} outputs) rather than multiple-choice classification, so gains there reflect transfer across task formats, not merely query distributions. Second, we evaluate on MMSI-Bench and MindCube below, both of which draw images from visual domains \emph{categorically distinct} from the natural-image scenes (MS-COCO, Visual Genome, ADE20K) used for calibration, including autonomous driving (Waymo, nuScenes), 3D indoor scenes (ScanNet, Matterport3D), and egocentric video (Ego4D). Consistent gains across this domain shift indicate that the calibrated reliability profiles generalize beyond the specific images seen during calibration.

\vspace{1mm}
\noindent\textbf{MMSI-Bench.}
We evaluate \textsc{SpatiO} on MMSI-Bench~\cite{yang2025mmsi}, a benchmark not included during TTO optimization, to test whether the unified taxonomy $\mathcal{C}$ transfers to unseen compositional reasoning tasks.

\begin{table}[H]
\centering
\caption{Zero-shot generalization of \textsc{SpatiO} on MMSI-Bench (optimized on
150 samples from 3DSRBench+STVQA+CV-Bench).}
\label{tab:supp_generalization}
\scriptsize
\setlength{\tabcolsep}{10pt}
\begin{tabular}{l ccccc}
\toprule
Method
& \makecell{MSR} 
& \makecell{Positional\\Relation} 
& \makecell{Motion} 
& \makecell{Attribute} 
& \cellcolor{gray!20}Overall \\
\midrule
\rowcolor{gray!15}
\multicolumn{6}{l}{\textit{Without LoRA training}} \\
Llava-4D            & \underline{24.7} & \underline{23.2} & 23.1 & 18.0 & \cellcolor{gray!20}23.2 \\
Qwen-3.0-VL-4B      & 24.3 & 22.2 & \underline{26.9} & \underline{23.3} & \cellcolor{gray!20}\underline{24.1} \\
Sa2VA               & 8.6 & 9.6 & 10.8 & 6.0 & \cellcolor{gray!20}8.7 \\
SpatialRGPT         & 22.4 & 15.0 & 17.4 & 3.2 & \cellcolor{gray!20}17.3 \\
SpatialReasoner     & 23.8 & 21.2 & 25.4 & 14.7 & \cellcolor{gray!20}22.1 \\
\midrule
\rowcolor{green!10}
\textsc{SpatiO} (Ours) & \textbf{54.3} & \textbf{41.0} & \textbf{42.1} & \textbf{37.0} & \cellcolor{green!20}\textbf{43.6} \\
\bottomrule
\end{tabular}
\end{table}

\textsc{SpatiO} achieves the highest overall accuracy in this fully zero-shot setting (+19.5\%p over the strongest baseline).
The largest gains occur on \textit{Positional Relationship} (+17.8\%) and \textit{Multi-Step Reasoning} (MSR, +29.6\%), while gains on \textit{Motion} (+15.2\%) and \textit{Attribute} (+13.7\%) are more moderate.

\noindent This pattern is explained by structural alignment between MMSI-Bench's categories and $\mathcal{C}$.
The \textit{Positional Relationship} category directly maps to \texttt{spatial\_relation} in $\mathcal{C}$, as both probe relative object placement (\eg, left/right, above/below).
The \textit{MSR} category maps to \texttt{distance\_depth}: multi-step spatial queries typically chain metric 3D inferences, depth ordering, relative distance estimation, and occlusion resolution in sequence, which are the operations encoded under \texttt{distance\_depth} in $\mathcal{C}$.
\textit{Motion} and \textit{Attribute} rely on temporal dynamics and appearance cues outside the static 3D scope of $\mathcal{C}$, explaining their comparatively smaller but still positive gains.
This structural alignment confirms that $\mathcal{C}$ captures a general decomposition of spatial reasoning that transfers across distribution shifts without benchmark-specific recalibration.

\vspace{1mm}
\noindent\textbf{MindCube.}
\label{sec:supp_mindcube}
As a further zero-shot probe drawn from a visual domain disjoint from the calibration pool, we evaluate on MindCube~\cite{yin2025spatial}.

\begin{table}[h]
\centering
\caption{\textbf{Zero-shot generalization on MindCube}~\cite{yin2025spatial}. \textsc{SpatiO} is calibrated on the same 150-sample set (3DSRBench, STVQA-7k, CV-Bench); no MindCube samples are used.}
\label{tab:mindcube}
\footnotesize
\setlength{\tabcolsep}{10pt}
\begin{tabular}{lc}
\toprule
\textbf{Method} & \textbf{Acc.\ (\%)} \\
\midrule
LLaVA-4D        & 31.43 \\
Qwen-3.0-VL-4B  & 30.26 \\
Sa2VA           & 42.72 \\
SpatialRGPT     & 34.51 \\
SpatialReasoner & 32.53 \\
\midrule
\rowcolor{green!10}
\textsc{SpatiO} (Ours) & \textbf{44.39} \\
\bottomrule
\vspace{2mm}
\end{tabular}
\end{table}

\vspace{2mm}
\textsc{SpatiO} again achieves the best overall accuracy (44.39\%), a +1.67\%p improvement over the strongest single-model baseline (Sa2VA, 42.72\%). The smaller margin relative to MMSI-Bench or Omni3D-Bench is consistent with MindCube's images being visually closer to the natural-image distribution underlying our specialist agents' pretraining, leaving less headroom for heterogeneous orchestration to exploit decorrelated failure modes; nonetheless, \textsc{SpatiO} remains the top-performing method, supporting the conclusion that the calibrated trust profiles do not merely overfit the three calibration benchmarks.

\vspace{3mm}

\subsection{Inference Latency Breakdown}
\label{sec:single_vs_spatio}

\noindent\textbf{Comparison against single-model baselines.}
Tab.~\ref{tab:single_vs_spatio} directly compares \textsc{SpatiO} against each single-model baseline on accuracy, average per-query latency, throughput, and peak VRAM on 3DSRBench (single A100 80GB GPU). \textsc{SpatiO} improves accuracy over the strongest single-model baseline, SpatialReasoner, from 65.6\% to 84.7\% (+19.1\%p) while \emph{reducing} average latency from 25.8\,s to 19.8\,s. This is possible because the Head Agent dynamically selects the most reliable specialist combination per query category, invoking heavyweight specialists like SpatialReasoner only when the spatial task demands it, rather than for every query; throughput and VRAM are correspondingly higher than any individual specialist alone, reflecting the cost of running multiple specialists in parallel.
\vspace{1mm}

\vspace{1mm}
\noindent\textbf{Per-module breakdown.}
\begin{table}[H]
\centering
\vspace{1mm}
\caption{Per-module inference latency (seconds per query, averaged over the full dataset from STVQA-7k). Specialists run in parallel; end-to-end latency is the maximum
across the three selected specialists plus Head and Final Reasoning Agent.}
\label{tab:single_vs_spatio}

\resizebox{\textwidth}{!}{%
\setlength{\tabcolsep}{4pt}
\begin{tabular}{l ccccc c}
\toprule
Module
& \makecell{Spatial\\Relation}
& \makecell{Distance\\Depth}
& \makecell{Size}
& \makecell{Counting}
& \makecell{Orientation}
& \cellcolor{gray!20}Avg. \\
\midrule
Head Agent (Qwen3-VL-4B)
& 2.4 & 2.6 & 2.3 & 2.5 & 2.7 & \cellcolor{gray!20}2.5 \\
\midrule
Role 1 · Implicit Visual
& 14.2 & 15.8 & 12.1 & 17.3 & 20.6 & \cellcolor{gray!20}16.0 \\
Role 2 · Explicit 3D
& 2.8 & 8.4 & 2.1 & 2.3 & 7.9 & \cellcolor{gray!20}4.5 \\
Role 3 · Scene Graph
& 9.3 & 5.2 & 4.8 & 8.6 & 4.1 & \cellcolor{gray!20}6.4 \\
\midrule
Final Reasoning Agent (DSR-7B)
& 4.8 & 5.1 & 3.9 & 4.2 & 4.5 & \cellcolor{gray!20}4.5 \\
\midrule
\rowcolor{green!10}
\textbf{End-to-end (\textsc{SpatiO})}
& 31.5 & 31.9 & 18.3 & 26.0 & 32.7 & \cellcolor{green!20}28.1 \\
\bottomrule
\end{tabular}
}
\end{table}

Tab.~\ref{tab:single_vs_spatio} reports per-module and end-to-end wall-clock latency (seconds per query, averaged over 200 CV-Bench samples) on 4$\times$A100 80GB GPUs.
Latency varies by category: from 18.3\,s for \textit{size} queries, where no deep depth pipeline or scene graph is required, to 32.7\,s for \textit{orientation} queries, where Role~2 invokes DepthPro~\cite{bochkovskii2024depth} and SAM2~\cite{ravi2024sam} (7.9--8.4\,s per query).
For relation-centric categories (\textit{spatial\_relation}, \textit{counting}), Role~3's DINOv2-based scene graph construction~\cite{oquab2023dinov2} becomes the bottleneck (8.6--9.3\,s), yielding moderate end-to-end latency of 26.0--31.5\,s.
The resulting Pareto frontier is therefore more advantageous than the average wall-clock figure alone suggests.
Several engineering improvements, asynchronous execution, KV-cache sharing, and depth-map reuse across adjacent frames, remain unimplemented and are expected to reduce latency further~\cite{chu2024mobilevlm}.

\subsection{Comparison with Prior Multi-Agent Frameworks}
\label{sec:supp_novelty}

Reviewers asked us to locate \textsc{SpatiO}'s novelty more explicitly relative to MATE~\cite{algazinov2025mate}, MoA/MoSA~\cite{wang2024mixture,yang2025multi}, and VADAR~\cite{marsili2025visual}, beyond the prompt-mediated role injection observation in Sec.~\ref{sec:related}. While each individual TTO component (Beta--Bernoulli trust modeling, dual-EMA smoothing, sigmoid-normalized top-$k$ routing) is independently well established, Tab.~\ref{tab:novelty_comparison} shows that no prior framework combines \emph{architectural heterogeneity}, \emph{structured geometric evidence} (depth maps, scene graphs), and \emph{per-(agent, role, category) trust estimation} in a single black-box-compatible, parameter-free system. MATE's Beta--Bernoulli trust model is restricted to abstract cooperative agents without grounded visual evidence or role-dependent behavior; MoA/MoSA aggregate heterogeneous LLM outputs but maintain only coarse, model-level (not role- or category-conditioned) trust; VADAR incorporates structured geometric evidence via program synthesis but relies on a single homogeneous backbone with rule-based aggregation. \textsc{SpatiO}'s contribution is therefore best understood as the \emph{composition} of these previously separate capabilities into a single spatially-grounded orchestration mechanism, rather than any single novel algorithmic primitive.

\begin{table}[h]
\centering
\caption{\textbf{Property-level comparison with prior multi-agent frameworks.} \textsc{SpatiO}'s key differentiator is per-role/category/agent trust estimation, combined with structured geometric evidence.}
\label{tab:novelty_comparison}
\footnotesize
\setlength{\tabcolsep}{5pt}
\resizebox{\linewidth}{!}{%
\begin{tabular}{lcccc}
\toprule
\textbf{Property} & \textbf{MATE} & \textbf{MoA/MoSA} & \textbf{VADAR} & \textbf{\textsc{SpatiO}} \\
\midrule
Heterogeneous agent pool                  & \xmark & \cmark & \xmark & \cmark \\
Per-role/category/agent trust estimation  & \xmark & \xmark & \xmark & \cmark \\
Structured geometric evidence             & \xmark & \xmark & \cmark & \cmark \\
Parameter-free adaptation                 & \cmark & \xmark & \xmark & \cmark \\
Vision-language input                     & \xmark & \xmark & \cmark & \cmark \\
Black-box model compatible                & \xmark & \cmark & \xmark & \cmark \\
\bottomrule
\end{tabular}%
}
\end{table}

\subsection{Head Agent Ablation: Full-Dataset Evaluation}
\label{sec:supp_head_ablation}

The Head Agent is responsible for routing each incoming query to the appropriate spatial category in $\mathcal{C}$, determining which per-category TTO weights govern downstream specialist selection.
Since routing errors propagate to all specialist activations, the choice of Head Agent model has a non-trivial impact on end-to-end performance.
We ablate five candidate Head Agents, Qwen3.0-VL-4B (default), Qwen3.0-VL-8B, Qwen3.0-VL-30B-A3B-Instruct (MoE), InternVL2-8B, and LLaVA-NeXT-7B, while holding all other components fixed.
Tables~\ref{tab:supp_head_cvbench} and~\ref{tab:supp_head_3dsrbench} report results on the full CV-Bench and full 3DSRBench datasets, respectively.

\vspace{1mm}
\begin{table}[h]
\centering
\caption{%
  \textbf{\textsc{SpatiO} Head Agent ablation on full CV-Bench (2,640 samples).}
  All specialist and reasoner agents are held fixed; only the Head Agent routing model varies.
}
\label{tab:supp_head_cvbench}
\setlength{\tabcolsep}{6pt}
\renewcommand{\arraystretch}{1.08}
\resizebox{\linewidth}{!}{
\begin{tabular}{l ccccc}
\toprule
{\textbf{Head Agent}} &
  \multicolumn{5}{c}{\textbf{CV-Bench-full}} \\
\cmidrule(lr){2-6}
 & \textbf{Count} & \textbf{Relation} & \textbf{Depth} & \textbf{Distance} & \textbf{Overall} \\
\midrule
Qwen3.0-VL-4B                      & 69.1 & 91.9 & 96.1 & 92.7 & \cellcolor{gray!20}87.47 \\
Qwen3.0-VL-8B                      & \textbf{71.4} & 90.1 & 96.0 & 92.3 & \cellcolor{gray!20}87.45 \\
Qwen3.0-VL-30B-A3B-Instruct (MoE)  & 70.0 & 93.2 & \textbf{96.4} & \textbf{92.9} & \cellcolor{gray!20}\textbf{88.13} \\
InternVL2-8B                       & 68.0 & \textbf{94.5} & 90.8 & 89.5 & \cellcolor{gray!20}85.70 \\
LLaVA-NeXT-7B                      & 70.5 & 88.4 & 84.1 & 92.2 & \cellcolor{gray!20}83.80 \\
\bottomrule
\end{tabular}%
}
\end{table}

\paragraph{Overall trends.}
All Head Agent variants preserve \textsc{SpatiO}'s competitive advantage over single-model baselines, confirming robustness to moderate routing model choices. However, routing quality introduces category-level variance that is non-uniform across benchmarks.

\noindent{\textbf{Qwen3.0-VL-4B vs.\ 8B on CV-Bench.}
The two variants achieve nearly identical overall accuracy (87.47 vs.\ 87.45), but differ in their category profiles. The 8B model leads on \textit{Count} (71.4 vs.\ 69.1): its stronger visual grounding may better disambiguate object instances. Conversely, the 4B model slightly outperforms on \textit{Relation} (91.9 vs.\ 90.1). This inversion suggests that scale does not uniformly improve routing precision, the 8B model appears to be slightly over-sensitive to counting cues, occasionally pulling borderline relation queries toward \texttt{counting}.

\noindent{\textbf{Qwen3.0-VL-30B-MoE: best overall, not best everywhere.}
The MoE variant achieves the highest overall accuracy on CV-Bench (88.13) and the second-highest on 3DSRBench (73.9), driven by strong \textit{Depth} (96.4) and \textit{Distance} (92.9) routing. However, its \textit{Count} accuracy (70.0) falls \emph{below} both Qwen 4B and 8B variants despite substantially more parameters. We attribute this to the MoE architecture's routing mechanism: mixture-of-experts models exhibit uneven expert specialization across token distributions~\cite{shazeer2017outrageously}, and counting queries, which rely on instance cardinality rather than spatial geometry, may not align well with any single expert's routing preference, resulting in inconsistent categorization of count queries even at large scale.

\noindent{\textbf{InternVL2-8B: strong on Relation.}
InternVL2-8B achieves the highest \textit{Relation} score on CV-Bench (94.5), consistent with its strong 2D grounding from interleaved image-text pretraining~\cite{chen2024expanding}. However, it underperforms markedly on \textit{Depth} (90.8) and \textit{Distance} (89.5), yielding the second-lowest overall accuracy (85.7) on CV-Bench. We hypothesize that InternVL2-8B's strong 2D spatial priors cause it to conflate depth-based queries with 2D relational ones, routing them to \texttt{spatial\_relation} rather than \texttt{distance\_depth} and thereby misapplying the metric-3D TTO weights.

\noindent{\textbf{LLaVA-NeXT-7B: consistent underperformance.}
LLaVA-NeXT-7B yields the lowest overall accuracy on both benchmarks (83.80 on CV-Bench; 65.8 on 3DSRBench). Its most pronounced weakness is \textit{Depth} (84.1), a 12-point gap relative to the MoE variant, and all orientation-related 3DSRBench subcategories, where it trails Qwen3.0-VL-4B by 6--8 points. LLaVA-NeXT-7B's predominantly 2D training corpus makes it an unreliable router for queries that hinge on perspective-based or metric 3D cues.

\vspace{2mm}
\begin{table}[h]
\centering
\caption{%
  \textbf{\textsc{SpatiO} Head Agent ablation on full 3DSRBench across all 12 subcategories.}
}
\label{tab:supp_head_3dsrbench}
\setlength{\tabcolsep}{2.8pt}
\renewcommand{\arraystretch}{1.08}
\resizebox{\linewidth}{!}{%
\begin{tabular}{l ccccccccccccc}
\toprule
{\textbf{Head Agent}} &
  \multicolumn{13}{c}{\textbf{3DSRBench}} \\
\cmidrule(lr){2-14}
 & \makecell{Loc.\\above} & \makecell{Height\\higher} & \makecell{Loc. closer\\to cam.} &
   \makecell{Multi-obj\\closer to} & \makecell{Orient.\\on left} & \makecell{Multi-obj\\facing} &
   \makecell{Multi-obj\\same dir.} & \makecell{Orient. in\\front of} &
   \makecell{Multi-obj VP\\toward obj.} & \makecell{Orient.\\VP} &
   \makecell{Loc.\\next to} & \makecell{Multi-obj\\parallel} & \textbf{Overall} \\
\midrule
Qwen3.0-VL-4B
  & 75.2 & 65.8 & 87.3 & 76.5 & 59.4 & 49.9 & 63.7 & 76.1 & 36.8 & 44.6 & 81.5 & 77.2 & \cellcolor{gray!20}72.4 \\
Qwen3.0-VL-8B
  & \textbf{78.1} & 68.2 & \textbf{89.0} & \textbf{79.3} & 61.2 & 51.4 & 65.5 & 77.9 & 34.1 & 46.2 & \textbf{83.8} & \textbf{79.0} & \cellcolor{gray!20}\textbf{74.5} \\
Qwen3.0-VL-30B-A3B-Instruct (MoE)
  & 76.4 & \textbf{70.1} & 87.8 & 77.6 & \textbf{63.9} & \textbf{54.2} & \textbf{67.8} & \textbf{80.3} & \textbf{39.5} & \textbf{49.3} & 82.1 & 77.5 & \cellcolor{gray!20}73.9 \\
InternVL2-8B
  & 71.8 & 62.4 & 85.1 & 74.2 & 57.3 & 47.1 & 60.4 & 73.5 & 37.6 & 43.1 & 79.2 & 74.8 & \cellcolor{gray!20}70.5 \\
LLaVA-NeXT-7B
  & 69.9 & 61.2 & 84.0 & 72.4 & 54.7 & 46.3 & 59.8 & 71.5 & 38.0 & 41.4 & 78.2 & 72.9 & \cellcolor{gray!20}68.4 \\
\bottomrule
\end{tabular}%
}
\end{table}

\paragraph{3DSRBench: category-level asymmetries.}
Tab.~\ref{tab:supp_head_3dsrbench} reveals asymmetries not apparent in overall accuracy.
The MoE variant excels on the hardest subcategories, \textit{Multi-obj VP toward object} (39.5), \textit{Orientation VP} (49.3), \textit{Multi-obj same direction} (67.8), which require resolving fine-grained 3D angular relationships: the MoE's diverse expert pool is better at disambiguating geometrically complex routing decisions.
In contrast, Qwen3.0-VL-8B leads on spatially simpler subcategories: \textit{Location above} (78.1), \textit{Location closer to camera} (89.0), and \textit{Location next to} (83.8), where routing involves straightforward depth-ordering or 2D proximity predicates.

Critically, the MoE model's \textit{Multi-obj VP toward object} score (39.5) is the highest yet still below 40\%, underscoring that viewpoint-centric subcategories remain a fundamental challenge for all routing models. Low absolute scores across all Head Agent choices in these categories confirm that the bottleneck lies in the specialist agents' 3D reasoning capacity, not in routing accuracy per se.

\paragraph{Default Head Agent choice.}
We adopt Qwen3.0-VL-4B as the default Head Agent.
While the MoE variant achieves marginally higher overall accuracy on CV-Bench (+0.66\%), the 4B model is 4$\times$ faster at routing (0.9\,s vs.\ 3.8\,s per query) and matches or exceeds the MoE on CV-Bench's high-frequency categories (\textit{Count}, \textit{Relation}).
This represents a favorable latency--accuracy trade-off for the Head Agent role, where precision on common query types matters more than marginal gains on the hardest orientation subcategories.

\vspace{2mm}


\section{Limitations and Future Work}
\label{sec:supp_limitations_futureworks}

\paragraph{Taxonomy and agent pool scalability.}
The current framework defines five spatial categories and maintains a fixed pool of five specialist agents. Expanding $\mathcal{C}$ to cover finer-grained predicates (affordance reasoning, egocentric navigation, temporal spatial change) and incorporating additional specialists may improve generalization.

\paragraph{Latency-aware orchestration.}
End-to-end latency is governed by the slowest active specialist. Future work could investigate adaptive early-exit mechanisms or dynamic specialist pruning conditioned on query difficulty. Engineering improvements, asynchronous specialist execution, KV-cache sharing, depth-map reuse across adjacent video frames, remain unimplemented and are expected to substantially reduce the latency gap~\cite{chu2024mobilevlm}.

\paragraph{Optimization data quality.}
Cross-benchmark experiments (Section~\ref{sec:supp_cross}) show that the choice of optimization source influences transfer performance. Purpose-built datasets that maximally cover the spatial predicate space of $\mathcal{C}$, or automatically synthesized hard cases for under-covered categories, could further strengthen TTO convergence and cross-benchmark robustness.

\paragraph{Extension to video and embodied settings.}
\textsc{SpatiO} currently operates on static images. Extension to video or embodied scenarios would require temporal modeling, depth-map reuse across frames, and adaptation of the TTO trust estimator to non-stationary query streams.

\vspace{3mm}

\section{Qualitative Results}
\label{sec:supp_qualitative}

We present qualitative examples illustrating the full pipeline outputs of \textsc{SpatiO} across representative spatial reasoning categories. All outputs shown are verbatim model responses from each specialist and the final reasoning agent, providing full transparency over the reasoning process.

A key aspect of \textsc{SpatiO}'s architecture is that all structured 3D data, depth maps, point cloud centroids, surface normals, inter-object distances, and scene graph edges, are passed to agents in serialised text form (see prompt templates in figures~\ref{fig:role2_tool}--\ref{fig:role3_tool}). While this representation is what the agents actually reason over, the qualitative figures additionally provide \textit{visual renderings} of the underlying 3D data to aid human interpretation: Role~2's tool outputs are rendered as 3D representations, and Role~3's scene graph is displayed as a directed graph diagram with labelled edges.
These visualisations are post-hoc and for illustration purposes only, they are not part of the agent's input.

\vspace{3mm}

\section{Prompt Design Rationale}
\label{sec:supp_prompt_rationale}

The prompt design reflects the insight that spatial reasoning is best accomplished through a set of complementary strategies, each attuned to different query types and evidence. Rather than forcing a single unified approach, our pipeline cultivates epistemic diversity: each agent contributes a distinct perspective, and their agreements or disagreements guide arbitration.

\vspace{-1mm}

\paragraph{Architectural philosophy.} The system is built as a layered evidence hierarchy. Role1 delivers rapid, lightweight reasoning from 2D pictorial cues; Role2 provides precise, tool-dependent 3D metrics; Role~3 enables expressive multi-object reasoning via scene graphs. This structure determines how the Final Reasoning Agent arbitrates, with TTO weights indexed by both role and query category.

\vspace{2mm}
\noindent\textbf{Head Agent (Fig.~\ref{fig:head_prompt}).}
The Head Agent functions as a zero-latency query router, committing to a single label before any specialist receives the query, so that any misclassification affects all subsequent reasoning. Its prompt encodes two main structural decisions. First, it enforces a strict single-label output with no abstention, preventing soft routing that would otherwise activate all specialists by default. Second, it introduces explicit disambiguation rules for semantically adjacent categories—for example, counting always overrides spatial context, so queries like “how many objects are to the left” are routed to counting rather than spatial relation, and near-synonyms such as “taller” and “higher” are assigned by geometry. These distinctions directly determine which specialist’s TTO weights are used during inference. The routing decision is intentionally irreversible, ensuring all specialists are conditioned on the same query framing without any option for downstream rerouting.

\vspace{2mm}
\noindent\textbf{Role 1 -- Implicit Visual Reasoning (Fig.~\ref{fig:role1_prompt}).}
Role~1 is designed to be maximally lightweight, relying solely on a
prioritised hierarchy of pictorial depth cues (occlusion $\succ$ relative size $\succ$ height-in-image $\succ$ familiar size) without any structured 3D input. Many spatial queries, especially left/right relations and vertical orderings,  are resolvable from 2D layout alone, making heavier pipelines unnecessary. Moreover, Role~1 serves as a consistency witness: agreement with Role~2 provides convergent evidence across modalities, while disagreement triggers principled arbitration by the Final Reasoning Agent rather than simple tie-breaking. The prompt encodes cue priority as an
explicit decision rule, applied through a mandatory decompose--anchor--resolve pipeline.

\vspace{2mm}
\noindent\textbf{Role 2 -- Explicit 3D Reconstruction (Figs.~\ref{fig:role2_prompt}--\ref{fig:role2_tool}).}
Role~2 is the \emph{metric anchor} of the pipeline. Its prompt is structured around an explicit trust ordering over tool outputs: depth maps from DepthPro are treated as the primary signal, instance masks and centroids from SAM2 as the secondary signal, and any implicit visual inference as a fallback of last resort. This ordering is not merely a preference, it is enforced as a hard constraint in the prompt, preventing the agent from rationalising a pictorial cue when a depth map is available. The structural consequence is that Role~2's outputs are \emph{quantitatively grounded} in a way that Roles~1 and~3 are not, which is why the Final Reasoning Agent assigns it dominant weight for \texttt{distance\_depth} and \texttt{orientation} queries where 2D layout cues are systematically unreliable.

\vspace{2mm}
\noindent\textbf{Role 3 -- Scene Graph Construction (Figs.~\ref{fig:role3_prompt}--\ref{fig:role3_tool}).}
Role~3 specializes in relational reasoning, using structured traversal of scene graphs and introducing enhancements such as inverse-edge lookup and pixel-level consistency checks. This makes it especially effective for multi-object and compositional queries. This design makes Role~3 most decisive for \texttt{spatial\_relation} and multi-object queries, where the relational predicate (e.g., ``between'', ``surrounding'', ``adjacent to'') is not recoverable from depth values alone and requires the compositional structure that a graph naturally encodes.

\vspace{2mm}
\noindent\textbf{Final Reasoning Agent (Fig.~\ref{fig:reasoning_prompt}).}
The Final Reasoning Agent differs from a majority voter or confidence-weighted ensemble. Its prompt uses a \emph{role-relevance priority table}, based on TTO weight profiles $w_{i,k,c}^{(t)}$, to indicate which specialist is most reliable for each query type. Rather than simply following these weights, the agent uses them to guide reasoning, but is also prompted to flag cases when a high-weight specialist’s output is inconsistent or conflicts with clear evidence from another role. 

\clearpage
\begin{figure}[t]
  \centering
  \includegraphics[width=\linewidth]{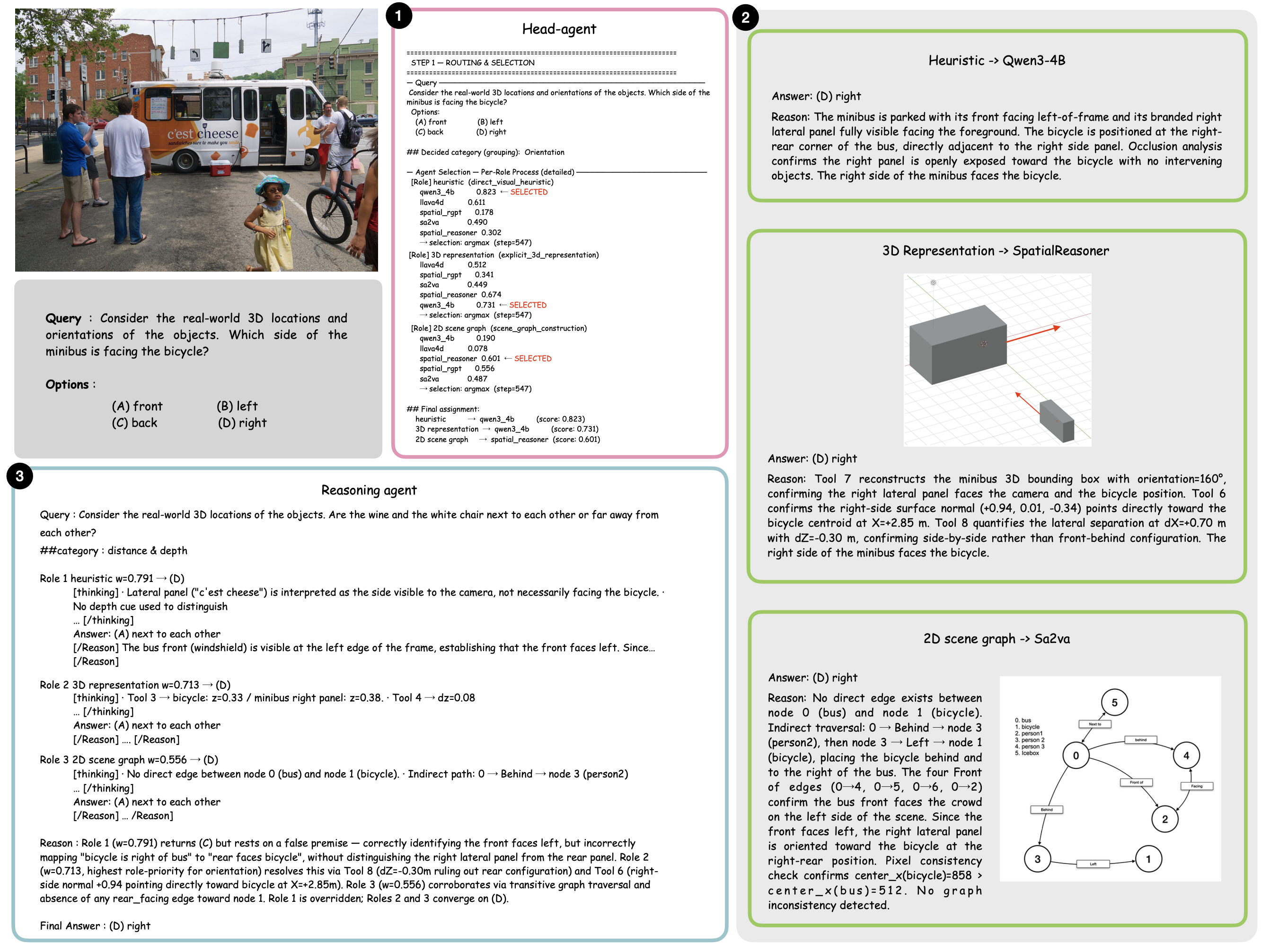}
  \\[2pt]
  \includegraphics[width=\linewidth]{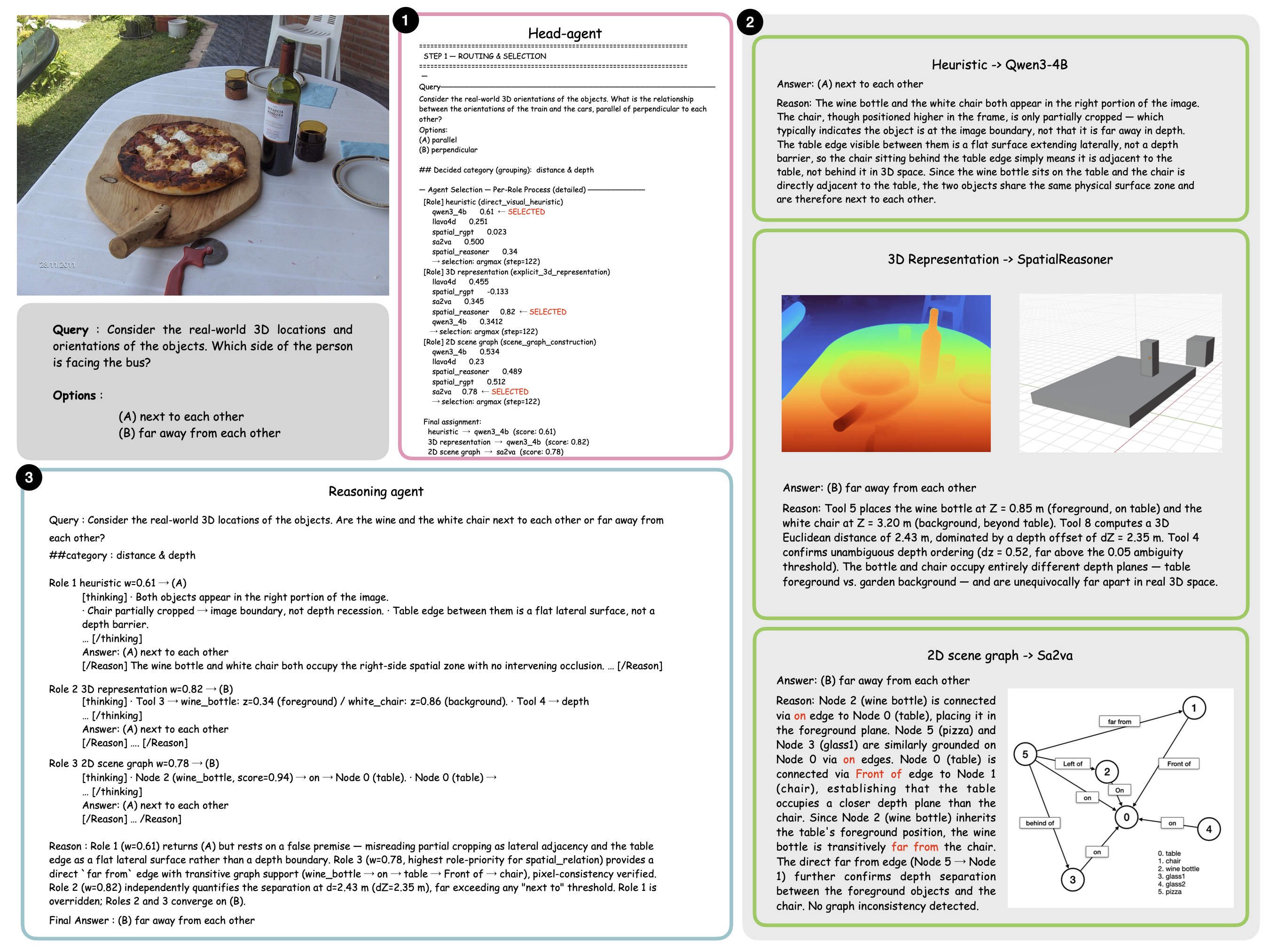}
  \caption{%
    \textbf{\textsc{SpatiO} pipeline output on 3DSRBench during TTO.}
    3D representations and scene graphs (SG) are post-hoc visualizations provided for interpretability only.
  }
  \label{fig:qualitative}
\end{figure}

\clearpage
\begin{figure}[!p]
  \centering \vspace*{\fill}
\begin{lstlisting}[style=PS]
# SYSTEM PROMPT | Head Agent -- Spatial Query Classifier
# -------------------------------------------------------
# Agent: Qwen3-VL-4B  |  Task: classify query into one of 5 unified spatial categories
# Output feeds FINE_TO_UNIFIED mapping -> Score Map -> specialist routing  (Sec. 4.3)

You are the Head Agent of SpatiO. Classify the spatial query into exactly ONE category
below, based on the primary geometric predicate the question asks to resolve.

Internally analyse before answering (do NOT output this):
  (1) Primary spatial predicate?  e.g. vertical stacking | lateral adjacency | depth | cardinality
  (2) Multiple predicates present? If so, which is the main ask?
  (3) Single object, or relationship between multiple objects?
  (4) Reference frame: camera/viewer, or another scene object?

## CATEGORIES
  spatial_relation   WHERE one object is relative to another in 3D space: above/below,
                     next to, between, on top of, at a higher elevation.
  distance_depth     HOW FAR from the camera/viewer or a third reference object.
  size               HOW BIG -- comparing physical scale, dimension, or extent.
  orientation        WHICH WAY -- facing direction, left/right, front/behind,
                     parallel or perpendicular arrangement relative to viewpoint.
  counting           HOW MANY -- cardinality of a set of objects in the scene.

## RULES
  1. HOW MANY / COUNT  ->  counting  (regardless of any spatial context in the question)
  2. Depth / proximity from camera  ->  distance_depth  (NOT spatial_relation)
  3. "taller" (physical size) -> size  |  "higher" (3D elevation) -> spatial_relation
  4. Facing direction, left/right, front/behind, angular arrangement  ->  orientation
  5. Classify by the predicate DIRECTLY ASKED, not by surrounding spatial context clauses.

## DO NOT  |  Output reasoning  |  Output anything except the category name  |  Invent categories

## INPUT / OUTPUT
  Question : {query}
  Output   : Respond with ONLY the category name. Nothing else.
\end{lstlisting}
  \caption{System prompt for the Head Agent (Qwen3-VL-4B). The routing decision is irreversible; disambiguation precision directly affects TTO weight selection for the entire pipeline.}
  \label{fig:head_prompt}
  \vspace*{\fill}
\end{figure}

\clearpage
\begin{figure}[!p]
  \centering \vspace*{\fill}
\begin{lstlisting}[style=PS]
# SYSTEM PROMPT | Role 1 -- Implicit Visual Reasoning Agent
# -----------------------------------------------------------------
# Assigned to: agents with dense 2D visual grounding  (Sec. 4.2)
# Strategy: pictorial depth cues only -- no explicit 3D reconstruction

You answer spatial reasoning questions using pictorial depth cues -- occlusion,
relative size, height in image, familiar size -- WITHOUT constructing explicit 3D models.
Always anchor reasoning to a reference object.

Internally select protocol (do NOT output):  HOW MANY/COUNT -> COUNT PROTOCOL  |  else -> SPATIAL PROTOCOL

## PICTORIAL CUES  (apply in priority order)
  occlusion       A hides B -> A is closer
  relative_size   Larger apparent size -> closer
  height_in_image Lower in frame -> closer
  familiar_size   Use known real-world size

## COUNT PROTOCOL  (HOW MANY / COUNT)
  1. Unit definition   ONE instance first. Multiple parts of same object = 1.
  2. Systematic scan   Top-left -> centre -> bottom-right -> edges.
  3. Occlusion rule    Partially visible distinct instance = 1.
  4. Semantic match    Broad definitions.  e.g. countertop as table -> include
  5. Re-check          List each instance. Avoid double-counting.

## SPATIAL PROTOCOL  (position / depth / distance / orientation)
  Step 1  Decompose    Atomic sub-questions.
  Step 2  Anchor       Reference object; describe its image position.
  Step 3  Cue+Resolve  Apply cues in order; state which cue supports conclusion.
                       Priority: occlusion > relative_size > height_in_image > familiar_size

## INPUT / OUTPUT
  Question : {query}
  Output   : Answer first, then justification <= 150 words.
             Format:  <Answer>  /  Reason: <justification>
\end{lstlisting}
  \caption{System prompt for Role~1: Implicit Visual Reasoning Agent. Agreement with Role~2 provides convergent evidence; disagreement triggers principled arbitration by the Final Reasoning Agent.}
  \label{fig:role1_prompt}
  \vspace*{\fill}
\end{figure}

\clearpage
\begin{figure}[!p]
  \centering \vspace*{\fill}
\begin{lstlisting}[style=PS]
# SYSTEM PROMPT | Role 2 -- Explicit 3D Reconstruction Agent
# -----------------------------------------------------------
# Assigned to: agents with explicit depth-aware architectures  (Sec. 4.2)
# Tools: DepthPro [1]  |  SAM2 [2]  |  SpatialVLM pipeline [3]

You answer spatial reasoning questions by interpreting structured 3D representations
computed from the input image. Do NOT rely on pictorial heuristics -- trust the tool outputs.

## TOOL SUITE
  depth_map       Dense per-pixel metric depth (DepthPro). Normalised z in [0,1], 0=closest.
  instance_mask   Per-object pixel mask and centroid (SAM2).
  point_cloud     3D centroids lifted from depth+mask (SpatialVLM). Centroid (X,Y,Z) + extent.
  surface_normal  Per-region normal vectors.
  bbox_3d         3D bounding box: centroid, dims (WxHxD m), orientation angle.

## 3D PROTOCOL  (depth / position / distance / orientation / size)
  Step 1  Decompose      Atomic sub-questions.
  Step 2  Select tools   depth ordering -> Tools 3,4  |  3D position -> Tool 5
                         inter-object distance -> Tool 8  |  size -> Tool 7  |  orientation -> Tools 6,7
  Step 3  Resolve        State numerical evidence; cite supporting tool.
                         Trust: point_cloud > depth_map > surface_normal > instance_mask

## INPUT / OUTPUT
  Question : {query}
  Output   : Answer first, then justification <= 150 words citing specific tool values.
             Format:  <Answer>  /  Reason: <justification with tool references>
\end{lstlisting}
  \caption{System prompt for Role~2: Explicit 3D Reconstruction Agent. The hard constraint against pictorial heuristics ensures quantitatively grounded outputs, justifying dominant TTO weight for \texttt{distance\_depth} and \texttt{orientation}.}
  \label{fig:role2_prompt}
  \vspace*{\fill}
\end{figure}

\clearpage
\begin{figure}[!p]
  \centering \vspace*{\fill}
\begin{lstlisting}[style=PS]
# TOOL OUTPUT TEMPLATE | Role 2 -- Explicit 3D Reconstruction Agent
# ------------------------------------------------------------------
## Tool 1.  Depth Map Grid  (3x3, DepthPro, normalised; 0=closest, 1=farthest)
  top-left:{tl}  top-center:{tc}  top-right:{tr}
  mid-left:{ml}  center:{cc}      mid-right:{mr}
  bot-left:{bl}  bot-center:{bc}  bot-right:{br}

## Tool 2.  Instance Masks & 2D Centroids  (SAM2)
  {obj_1}: centroid=({cx},{cy}px)  bbox=[{x1},{y1},{x2},{y2}]  area={area}px2

## Tool 3.  Object Depth Values  (DepthPro, normalised z)
  1. {obj_1}: z={val}  [CLOSEST]     2. {obj_2}: z={val}     3. {obj_3}: z={val}  [FARTHEST]

## Tool 4.  Depth Ordering       {obj_A} (z={za}) -> {obj_B} (z={zb})  [dz={delta}]
## Tool 5.  3D Point Cloud       {obj_1}: centroid=(X={x}, Y={y}, Z={z}m)  extent=(W={w},H={h},D={d}m)
## Tool 6.  Surface Normals      {obj_1}: normal=({nx},{ny},{nz})  facing={front/up/left/right/away}
## Tool 7.  3D Bounding Boxes    {obj_1}: centroid=(X,Y,Z)  dims=(WxHxD m)  orientation={angle}deg
## Tool 8.  Inter-Object Dist.   {obj_A} <-> {obj_B}: d={dist}m  (dX={dx}, dY={dy}, dZ={dz})
## Tool 9.  Instance Count       {object_type}: {count}

## Interpretation Guide
  z (norm.)  0=closest, 1=farthest.  dz < 0.05 -> ambiguous; verify with Tool 5.
  Trust      point_cloud > depth_map > surface_normal > instance_mask > pictorial
\end{lstlisting}
  \caption{Tool output template for Role~2. Tools~1--4: DepthPro depth evidence; Tool~2: SAM2 instance localisation; Tools~5--8: SpatialVLM metric 3D geometry.}
  \label{fig:role2_tool}
  \vspace*{\fill}
\end{figure}

\clearpage
\begin{figure}[!p]
  \centering \vspace*{\fill}
\begin{lstlisting}[style=PS]
# SYSTEM PROMPT | Role 3 -- Scene Graph Construction Agent
# ---------------------------------------------------------
# Assigned to: agents with relational graph reasoning capability  (Sec. 4.2)
# Graph extracted via: DINOv2 [4] with relational edge labelling

You answer spatial reasoning questions by combining: (1) Image, (2) Query, (3) Scene graph.
Graph is PRIMARY structured source. Always cross-check with the image.
If graph contradicts image -> prioritise image; state "Graph inconsistency detected."

## GRAPH PROTOCOL  (relation / position / orientation / depth)
  Step 1  Parse query    (a) subject  (b) reference object  (c) relation type
  Step 2  Locate nodes   Label match; multiple same-label -> pick highest score.
  Step 3  Traverse       Direct:  edge(subject=A, relation=R, object=B) -> confirms R(A,B)
                         Inverse: edge(subject=B, relation=R', object=A) -> confirms R(A,B)
                         Pairs:   above<->below  |  left_of<->right_of  |  in_front_of<->behind
  Step 4  Cross-check    Verify result vs image layout. If contradicted -> trust image.
  Step 5  Map to option  Answer node label -> correct option (A)/(B)/(C)/(D).

## FALLBACK  |  Empty/failed -> image only.  |  Edge missing -> check inverse.
              |  Graph contradicts image -> trust image; "Graph inconsistency detected."

## INPUT / OUTPUT
  Question : {query}
  Output   : Answer first, then justification <= 150 words citing edges and node IDs.
             Format:  <Answer>  /  Reason: <justification with graph references>
\end{lstlisting}
  \caption{System prompt for Role~3: Scene Graph Construction Agent. Inverse-edge lookup and pixel-level consistency checks make it most decisive for \texttt{spatial\_relation} queries.}
  \label{fig:role3_prompt}
  \vspace*{\fill}
\end{figure}

\clearpage
\begin{figure}[!p]
  \centering \vspace*{\fill}
\begin{lstlisting}[style=PS]
# TOOL OUTPUT TEMPLATE | Role 3 -- Scene Graph Construction Agent
# ---------------------------------------------------------------
## Scene Graph (JSON)
{
  "nodes": [
    {"id":"1","label":"{obj_1}","bbox":[{x1},{y1},{x2},{y2}],"center":[{cx},{cy}],"score":{conf}},
    {"id":"2","label":"{obj_2}","bbox":[{x1},{y1},{x2},{y2}],"center":[{cx},{cy}],"score":{conf}}
  ],
  "edges": [
    {"subject":"{id}","relation":"{rel}","object":"{id}"}
  ]
}
## Edge relations: above|below|left_of|right_of|in_front_of|behind|overlaps|closer_to|farther_from
## Consistency Check
  above/below      center_y(subject) < center_y(object)   [Y: 0=top]
  left_of/right_of center_x(subject) < center_x(object)   [X: 0=left]
  Coordinates contradict edge -> flag; reason from image.
\end{lstlisting}
  \caption{Tool output template for Role~3. Nodes: DINOv2 detection with confidence; edges: directed spatial predicates. Consistency Check enables pixel-level cross-validation against image layout.}
  \label{fig:role3_tool}
  \vspace*{\fill}
\end{figure}

\clearpage
\begin{figure}[!p]
  \centering \vspace*{\fill}
\begin{lstlisting}[style=PS]
# SYSTEM PROMPT | Final Reasoning Agent -- Weighted Evidence Integration
# -----------------------------------------------------------------------
# Agent: DeepSeek-VL-R1-7B  |  TTO weights w_{i,k,c}^(t) (Eq. 3)

You are the Final Reasoning Agent of SpatiO. Use TTO weight w to MODULATE -- not replace -- reasoning.

## STEP 2 -- EVALUATE AGENTS
  High w + relevant role -> strong signal.  Low w -> weak evidence regardless of answer.
  High-w agent with IRRELEVANT role: down-weight by relevance.

## STEP 3 -- WEIGHTED SYNTHESIS
  Disagreement -> favour role-matched agent citing most concrete data (z values, graph edges).
  Do NOT follow majority vote blindly. One high-w relevant agent can override two weaker ones.

## ROLE-RELEVANCE PRIORITY
  spatial_relation / size   scene_graph > implicit_visual > explicit_3d
  distance_depth            explicit_3d > implicit_visual > scene_graph
  orientation               explicit_3d > scene_graph     > implicit_visual
  counting                  scene_graph > explicit_3d     > implicit_visual

## OUTPUT FORMAT  (STRICT)
  multiple_choice   Answer: (A)/(B)/(C)/(D)
                    Reason: <2-5 sentences>
  open_ended        Answer: <value>
                    Reason: <2-5 sentences>
\end{lstlisting}
  \caption{System prompt for the Final Reasoning Agent (DeepSeek-VL-R1-7B). TTO weights act as a calibrated prior that modulates rather than replaces reasoning, preserving the ability to recover from systematic specialist errors. Role-priority table and open-ended aggregation rules cover benchmarks such as Omni3D-Bench~\cite{brazil2023omni3d} requiring direct numerical estimation.}
  \label{fig:reasoning_prompt}
  \vspace*{\fill}
\end{figure}

%
%

\end{document}